\documentclass{article} 
\usepackage{nips12submit_e,times}

\usepackage{graphicx} 
\usepackage{subfigure} 
\usepackage{array} 

\DeclareGraphicsExtensions{.eps,.eps.jpg.eps}

\usepackage{natbib}

\usepackage{algorithm}
\usepackage{algorithmic}

\usepackage{times,amsmath} 
\usepackage{psfrag}
\usepackage{stfloats}
\usepackage{multirow}
\usepackage{rotating}
\usepackage{amssymb}
\usepackage{paralist}
\usepackage{subfigure} 

\newcommand{\bmx}[0]{\begin{bmatrix}}
\newcommand{\emx}[0]{\end{bmatrix}}

\newcommand{\vect}[1]{\mathbf{#1}}
\newcommand{\vects}[1]{\boldsymbol{#1}}
\newcommand{\matr}[1]{\mathbf{#1}}

\newcommand{\vh}[0]{\vect{h}}
\newcommand{\vv}[0]{\vect{v}}
\newcommand{\vx}[0]{\vect{x}}

\newcommand{\vone}[0]{\vect{1}}
\newcommand{\mW}[0]{\matr{W}}

\newcommand{\mU}[0]{\matr{U}}

\newcommand{\mD}{\matr{D}}

\newcommand{\vmu}[0]{\vects{\mu}}

\newcommand{\TT}[0]{{\vects{\theta}}}

\newcommand{\LL}[0]{\mathcal{L}}

\newcommand{\E}[0]{\mathbb{E}}
\newcommand{\RR}[0]{\mathbb{R}}

\title{Boltzmann Machines and Denoising Autoencoders \\ for Image Denoising}

\author{
KyungHyun Cho \\
Aalto University School of Science \\
Department of Information and Computer Science \\
Espoo, Finland \\
\texttt{kyunghyun.cho@aalto.fi} \\
}

\nipsfinalcopy 

\begin{document}

\maketitle

\begin{abstract}
    Image denoising based on a probabilistic model of local
    image patches has been employed by various researchers,
    and recently a deep (denoising) autoencoder has been
    proposed by \citet{Burger2012} and \citet{Xie2012} as a
    good model for this.  In this paper, we propose that
    another popular family of models in the field of
    \textit{deep learning}, called Boltzmann machines, can
    perform image denoising as well as, or in certain cases
    of high level of noise, better than denoising
    autoencoders. We empirically evaluate the two models on
    three different sets of images with different types and
    levels of noise.  Throughout the experiments we also
    examine the effect of the depth of the models. The
    experiments confirmed our claim and revealed that the
    performance can be improved by adding more hidden
    layers, especially when the level of noise is high.
\end{abstract}

\section{Introduction}

Numerous approaches based on machine learning have been
proposed for image denoising tasks over time. A dominant
approach has been to perform denoising based on local
statistics over a whole image. For instance, a method
denoises each small image patch extracted from a whole noisy
image and reconstructs the clean image from the denoised
patches. Under this approach, it is possible to use raw
pixels of each image patch \citep[see,
e.g.][]{Hyvarinen1999a} or the representations in another
domain, for instance, in wavelet domain \citep[see,
e.g.][]{Portilla2003}.

In the case of using raw pixels, sparse coding has been a
method of choice. \citet{Hyvarinen1999a} proposed to use
independent component analysis (ICA) to estimate a
dictionary of sparse elements and compute the sparse code of
image patches. Subsequently, a shrinkage nonlinear function
is applied to the estimated sparse code elements to suppress
those elements with small absolute magnitude. This sparse
code elements are used to reconstruct a noise-free image
patch.  More recently, \citet{Elad2006} also showed that
sparse overcomplete representation is useful in denoising
images. Some researchers claimed that better
denoising performance can be achieved by using a variant of
sparse coding methods \citep[see, e.g.][]{Shang2005,Lu2011}.

In essence, these approaches build a probabilistic model of
natural image patches using a layer of sparse latent
variables. The posterior distribution of each noisy patch is
either exactly computed or estimated, and the noise-free
patch is reconstructed as an expectation of a conditional
distribution over the posterior distribution.

Based on this interpretation, some researchers have proposed
very recently to utilize a (probabilistic) model that has
more than one layers of latent variables for image
denoising.  \citet{Burger2012} showed that a deep
multi-layer perceptron that learns a mapping from a noisy
image patch to its corresponding clean version, can perform
as good as the state-of-the-art denoising methods.
Similarly,
\citet{Xie2012} proposed a variant of a stacked denoising
autoencoder \citep{Vincent2010} that is more effective in
image denoising. They were also able to show that the
denoising approach based on deep neural networks performed
as good as, or sometimes better than, the conventional
state-of-the-art methods.

Along this line of research, we aim to propose a yet another
type of deep neural networks for image denoising, in this
paper. A Gaussian-Bernoulli restricted Boltzmann machines
(GRBM) \citep{Hinton2006} and deep Boltzmann
machines (GDBM) \citep{Salakhutdinov2009a,Cho2011dlufl} are
empirically shown to perform well in image denoising,
compared to stacked denoising autoencoders. Furthermore, we
extensively evaluate the effect of the number of hidden
layers of both Boltzmann machine-based deep models and
autoencoder-based ones. The empirical evaluation is
conducted using different noise types and levels on three
different sets of images.

\section{Deep Neural Networks}

We start by briefly describing Boltzmann machines and
denoising autoencoders which have become increasingly
popular in the field of machine learning. 

\subsection{Boltzmann Machines}

Originally proposed in 1980s, a Boltzmann machine (BM)
\citep{Ackley1985} and especially its structural constrained
version, a restricted Boltzmann machine (RBM)
\citep{Smolensky1986} have become increasingly important in
machine learning since \citep{Hinton2006} showed that a
powerful deep neural network can be trained easily by
stacking RBMs on top of each other. More recently, another
variant of a BM, called a deep Boltzmann machine (DBM), has
been proposed and shown to outperform other conventional
machine learning methods in many tasks \citep[see,
e.g.][]{Salakhutdinov2009a}.

We first describe a Gaussian-Bernoulli DBM (GDBM) that has
$L$ layers of binary hidden units and a single layer of
Gaussian visible units. A GDBM is defined by its energy
function
\begin{align}
    \label{eq:energy_dbm}
    -E&(\vv, \vh \mid \TT) = 
    \sum_{i} -\frac{(v_i - b_i)^2}{2 \sigma^2} +
    \sum_{i,j} \frac{v_i}{\sigma^2} h_j^{(1)} w_{i,j} + 
    \sum_j h_j^{(1)} c_j^{(1)} +
    \nonumber\\
    &\sum_{l=2}^{L} \left( \sum_j 
    h_j^{(l)} c_j^{(l)} +
    \sum_{j,k} h_j^{(l)} h_k^{(l+1)}
    u_{j,k}^{(l)}
    \right),
\end{align}
where $\vv=\left[ v_i \right]_{i=1 \dots N_v}$ and
$\vh^{(l)}=\left[ h_j^{(l)} \right]_{j=1 \dots N_l}$ are
$N_v$ Gaussian visible units and $N_l$ binary hidden units
in the $l$-th hidden layer. $\mW = \left[ w_{i,j} \right]$
is the set of weights between the visible neurons and the
first layer hidden neurons, while $\mU^{(l)} = \left[
u_{j,k}^{(l)} \right]$ is the set of weights between the
$l$-th and $l+1$-th hidden neurons. $\sigma^2$ is the shared
variance of the conditional distribution of $v_i$ given the
hidden units.

With the energy function, a GDBM can assign a probability to
each state vector $\vx = [ \vv ; \vh^{(1)} ; \cdots ; \vh^{(L)} ]$
using a Boltzmann distribution:
\begin{align*}
    p(\vx \mid \TT) = \frac{1}{Z(\TT)} \exp \left\{
    -E(\vx\mid \TT) \right\}.
\end{align*}
Based on this property the parameters can be learned by
maximizing the log-likelihood $\LL = \sum_{n=1}^N \log
\sum_\vh p(\vv^{(n)}, \vh \mid \TT)$ given $N$ training
samples $\{ \vv^{(n)} \}_{n=1, \dots,N}$, where $\vh =
\left[ \vh^{(1)} ; \cdots ; \vh^{(L)} \right]$.


Although the update rules based on the gradients of the
log-likelihood function
are well defined, it is intractable to exactly compute them.
Hence, an approach that uses variational approximation
together with Markov chain Monte Carlo (MCMC) sampling was
proposed by \citet{Salakhutdinov2009a}. 

It has, however, been found that training a GDBM using this
approach starting from randomly initialized parameters is
not trivial
\citep{Salakhutdinov2009a,Desjardins2012,Cho2012dlufl}.
Hence, \citet{Salakhutdinov2009a} and \citet{Cho2012dlufl}
proposed pretraining algorithms that can initialize the
parameters of DBMs. In this paper, we use the pretraining
algorithm proposed by \citet{Cho2012dlufl}.

A Gaussian-Bernoulli RBM (GRBM) is a special case of a GDBM,
where the number of hidden layers is restricted to one,
$L=1$. Due to this restriction it is possible to compute the
posterior distribution over the hidden units conditioned on
the visible units exactly and tractably.  The conditional
probability of each hidden unit $h_j =
h_j^{(1)}$ is 
\begin{align} 
    \label{eq:rbm_cond} 
    p(h_j = 1 \mid \vv, \TT) = f\left( \sum_i w_{ij}
    \frac{v_i}{\sigma^2} + c_j\right).  
\end{align}

Hence, the positive part of the gradient, that needs to be
approximated with variational approach in the case of GDBMs,
can be computed exactly and efficiently. Only the negative
part, which is computed over the model distribution, still
relies on MCMC sampling, or more approximate methods such as
contrastive divergence (CD) \citep{Hinton2002}.

\subsection{Denoising Autoencoders}

A denoising autoencoder (DAE) is a special form of multi-layer
perceptron network with $2 L
- 1$ hidden layers and $L-1$ sets of tied weights. A DAE
tries to learn a network that reconstructs an input vector
optimally by minimizing the following cost function:
\begin{align}
    \sum_{n=1}^N \left\| \mW g^{(1)} \circ \cdots \circ
    g^{(L-1)} \circ f^{(L-1)} \circ \cdots \circ f^{(1)} \left(
    \eta(\vv^{(n)})\right) -
    \vv^{(n)} \right\|^2,
\end{align}
where 
\begin{align*}
    f^{(l)} = \phi( {\mW^{(l)}}^\top \vh^{(l-1)})\text{ and
    }
    g^{(l)} = \phi(\mW^{(l)} \vh^{(2L - l)})
\end{align*}
are encoding and decoding functions for $l$-th layer with a
component-wise nonlinearity function $\phi$. $\mW^{(l)}$ is
the weights between the $l$-th and $l+1$-th layers and is
shared by the encoder and decoder. For notational simplicity, we omit
biases to all units.

Unlike an ordinary autoencoder, a DAE explicitly sets some
of the components of an input vector randomly to zero during
learning via $\eta(\cdot)$ which explicitly adds noise to an
input vector. It is usual to combine two different types of
noise when using $\eta(\cdot)$, which are additive isotropic
Gaussian noise and masking noise \citep{Vincent2010}. The
first type adds a zero-mean Gaussian noise to each input
component, while the masking noise sets a set of randomly
chosen input components to zeros. Then, the DAE is trained
to \textit{denoise} the corrupted input.

Training a DAE is straightforward using backpropagation
algorithm which computes the gradient of the objective
function using a chain-rule and dynamic programming.
\citet{Vincent2010} proposed that training a deep DAE
becomes easier when the weights of a deep DAE are
initialized by greedily pretraining each layer of a deep DAE
as if it were a single-layer DAE. In the following
experiments section, we follow this approach to initialize
the weights and subsequently finetune the network with the
stochastic backpropagation.

\section{Image Denoising}

There are a number of ways to perform image denoising. In
this paper, we are interested in an approach that relies on
local statistics of an image.

As it has been mentioned earlier, a noisy large image can be
denoised by denoising small patches of the image and
combining them together. Let us define a set of $N$ binary
matrices $\mD_n \in \RR^{p \times d}$ that extract a set of
small image patches given a large, whole image $\vx \in
\RR^d$, where $d = w h c$ is the product of the width $w$,
the height $h$ and the number of color channels $c$ of the
image and $p$ is the size of image patches (e.g., $n=64$ if
an $8\times 8$ image patch).  Then, the denoised image is
constructed by
\begin{align}
    \label{eq:image_denoising}
    \tilde{\vx} = 
    \left(\sum_{n=1}^N \mD_n^\top r_{\TT}(\mD_n \vx)\right)
    \oslash \left(\sum_{n=1}^N \mD_n^\top \mD_n \vone\right) ,
\end{align}
where $\oslash$ is a element-wise division and $\vone$ is a
vector of ones. $r_{\TT} (\cdot)$ is an image denoising
function, parameterized by $\TT$, that denoises $N$ image
patches extracted from the input image $\vx$.

Eq.~\eqref{eq:image_denoising} essentially extracts and
denoises all possible image patches from the input image.
Then, it combines them by taking an average of those
overlapping pixels.

There are several flexibilities in constructing a matrix
$\mD$. The most obvious one is the size of an image patch.
Although there is no \textit{standard} approach, many
previous attempts tend to use patch sizes as small as $4
\times 4$ to $17 \times 17$. Another one, called a stride, is
the number of pixels between two consecutive patches. Taking
every possible patch is one option, while one may opt to
overlapping patches by only a few pixels, which would reduce
the computational complexity.

One of the popular choices for $r_{\TT}(\cdot)$ has been to
construct a probabilistic model with a set of latent
variables that describe natural image patches. For instance,
sparse coding which is, in essence, a probabilistic model
with a single layer of latent variables has been a common
choice.  \citet{Hyvarinen1999a} used ICA and a nonlinear
shrinkage function to compute the sparse code of an image
patch, while \citet{Elad2006} used K-SVD to build a sparse
code dictionary. 

Under this approach denoising can be considered as a
two-step reconstruction. Initially, the posterior
distribution over the latent variables is computed, or
estimated, given an image patch. Given the estimated
posterior distribution, the conditional distribution, or its
mean, over the visible units is computed and used as a
denoised image patch.

In the following subsections, we describe how
$r_{\TT}(\cdot)$ can be implemented in the cases of
Boltzmann machines and denoising autoencoders.

\subsection{Boltzmann machines}

We consider a BM with a set of Gaussian visible units $\vv$
that correspond to the pixels of an image patch and a set of
binary hidden units $\vh$. Then, the goal of denoising can
be written as
\begin{align}
    \label{bm:denoise_cond}
    p(\vv \mid \tilde{\vv}) = \sum_{\vh} p(\vv \mid
    \vh) p(\vh \mid \tilde{\vv}) = \E_{\vh \mid \tilde{\vv}}
    \left[ p(\vv \mid \vh)\right],
\end{align}
where $\tilde{\vv}$ is a noisy input patch. In other words,
we find a mean of the conditional distribution of the
visible units with respect to the posterior distribution
over the hidden units given the visible units fixed to the
corrupted input image patch.

However, since taking the expectation over the posterior
distribution is usually not tractable nor exactly
computable, it is often easier to approximate the quantity.
We approximate the marginal conditional distribution
\eqref{bm:denoise_cond} of $\vv$ given $\tilde{\vv}$ with
\begin{align*}
    p(\vv \mid \tilde{\vv}) &\approx p(\vv \mid \vh) Q(\vh)
    = p(\vv \mid \vh = \vmu),
\end{align*}
where $\vmu = \E_{Q(\vh)} \left[ \vh \right]$ and $Q(\vh)$
is a fully factorial distribution that approximates the
posterior $p(\vh \mid \tilde{\vv})$. It is usual to
use the mean of $\vv$ under $p(\vv \mid \tilde{\vv})$ as the
denoised, reconstructed patch.

Following this approach, given a noisy image patch
$\tilde{\vv}$ a GRBM reconstructs a noise-free patch by
\begin{align*}
    \hat{v}_i = \sum_{j=1}^{N_h} w_{ij} \E\left[\vh \mid
    \tilde{\vv}\right] + b_i,
\end{align*}
where $b_i$ is a bias to the $i$-th visible unit. The
conditional distribution over the hidden units can be
computed exactly from Eq.~\eqref{eq:rbm_cond}.

Unlike a GRBM, the posterior distribution of the hidden
units of a GDBM is neither tractably computable nor has an
analytical form. \citet{Salakhutdinov2009a} proposed to
utilize a variational approximation \citep{Neal1999} with a
fully-factored distribution $Q(\vh) = \prod_{l=1}^{L}
\prod_{j} \mu_j^{l}$, where the variational parameters
$\mu_j^{(l)}$'s can be found by the following simple
fixed-point update rule:
\begin{align}
    \label{eq:dbm_mf}
  \mu_j^{(l)} \leftarrow f \left( \sum_{i=1}^{N_{l-1}} \mu_i^{(l-1)}
    w_{ij}^{(l-1)} +\sum_{k=1}^{N_{l+1}} \mu_k^{(l+1)} w_{kj}^{(l)}
    +c_j^{(l)} \right),
\end{align}
where $f(x) = \frac{1}{1+\exp\left\{-x\right\}}$.

Once the variational parameters are converged, a GDBM
reconstructs a noise-free patch by
\begin{align*}
    \hat{v}_i = \sum_{j=1}^{N_l} w_{ij} \mu_j^{(1)} + b_i.
\end{align*}

The convergence of the variational parameters can take too
much time and may not be suitable in practice.  Hence, in
the experiments, we initialize the variational parameters by
feed-forward propagation using the doubled weights
\citep{Salakhutdinov2009a} and perform the fixed-point
update in Eq.~\eqref{eq:dbm_mf} for at most five iterations
only. This turned out to be a good enough compromise that
least sacrifices the performance while reducing the
computational complexity significantly.

\subsection{Denoising autoencoders}

An encoder part of a DAE can be considered as performing an
approximate inference of a fully-factorial posterior distribution
of top-layer hidden units, i.e. a bottleneck, given an input
image patch \citep{Vincent2010}. Hence, a similar approach 
to the one taken by BMs can be used for DAEs. 

Firstly, the variational parameters $\vmu^{(L)}$ of the
fully-factorial posterior distribution $Q(\vh^{(L)}) =
\prod_{j} \mu_j^{(L)}$ are computed by
\begin{align*}
    \vmu^{(L)} = f^{(L-1)} \circ \cdots \circ f^{(1)}
    \left( \tilde{\vv} \right).
\end{align*}

Then, the denoised image patch can be reconstructed by the
decoder part of the DAE. This can be done simply by
propagating the variational parameters through the decoding
nonlinearity functions $g^{(l)}$ such that
\begin{align*}
    \hat{\vv} = g^{(1)} \circ \cdots \circ g^{(L-1)}
    \left(\vmu^{(L)}\right).
\end{align*}

Recently, \citet{Burger2012} and \citet{Xie2012} tried a
deep DAE in this manner to perform image denoising. Both of
them reported that the denoising performance
achieved by DAEs is comparable, or sometimes favorable, to
other conventional image denoising methods such as BM3D
\citep{Dabov2007}, K-SVD \citep{Portilla2003} and Bayes
Least Squares-Gaussian Scale Mixture \citep{Elad2006}.

\section{Experiments}

In the experiments, we aim to empirically compare the two
dominant approaches of \textit{deep learning}, namely
Boltzmann machines and denoising autoencoders, in image
denoising tasks. 

There are several questions that are of interest to us:
\begin{enumerate}
    \item Does a model with more hidden layers perform
        better? 
    \item How well does a deep model generalize?
    \item Which family of deep neural networks is more
        suitable, Boltzmann machines or denoising
        autoencoders?
\end{enumerate}

In order to answer those questions, we vary the depth of the
models (the number of hidden layers), the level of noise
injection, the type of noise--either white Gaussian additive
noise or salt-and-pepper noise, and the size of image
patches. Also, as our interest lies in the generalization
capability of the models, we use a completely separate data
set for training the models and apply the trained models to
three distinct sets of images that have very different
properties.

\subsection{Datasets}

\begin{figure}[t]
    \centering
    \begin{minipage}{0.32\textwidth}
        \centering
        \includegraphics[width=0.48\columnwidth,clip=true,trim=120 45 110 40]{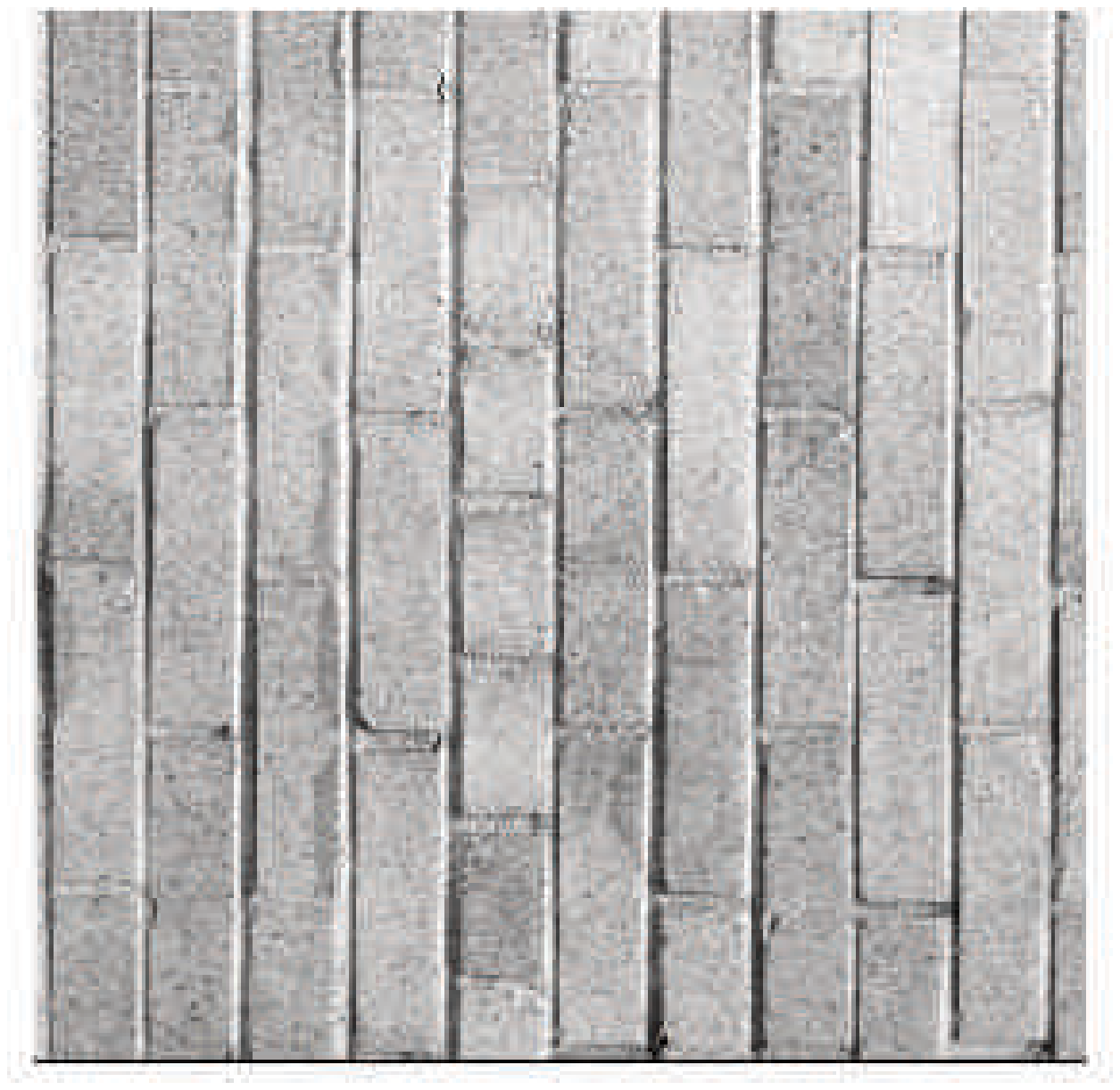}
        \includegraphics[width=0.48\columnwidth,clip=true,trim=120
        45 110 40]{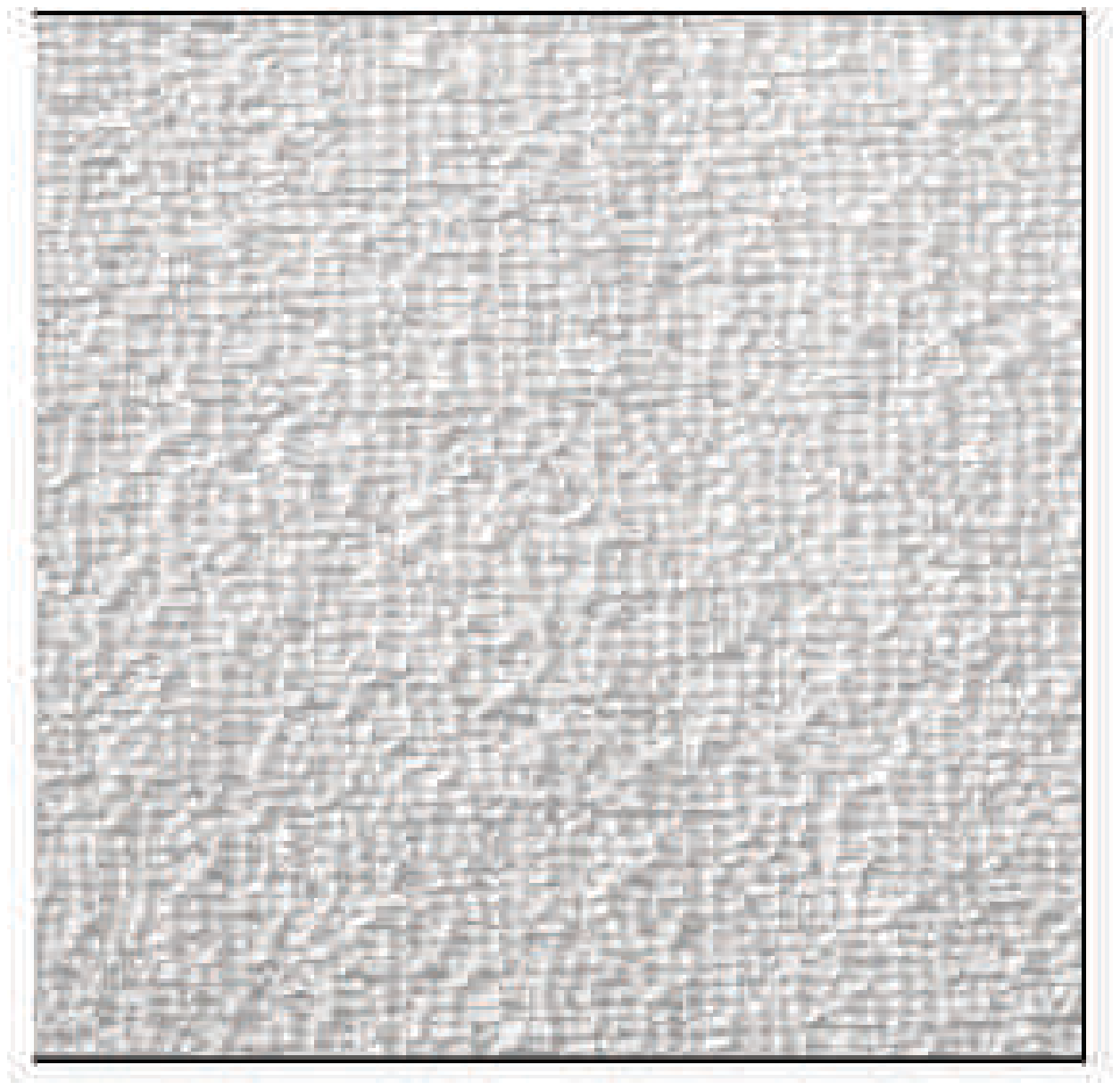}
        \\
        (a) Textures
    \end{minipage}
    \begin{minipage}{0.32\textwidth}
        \centering
        \includegraphics[width=0.48\columnwidth,clip=true,trim=120
        45 110 40]{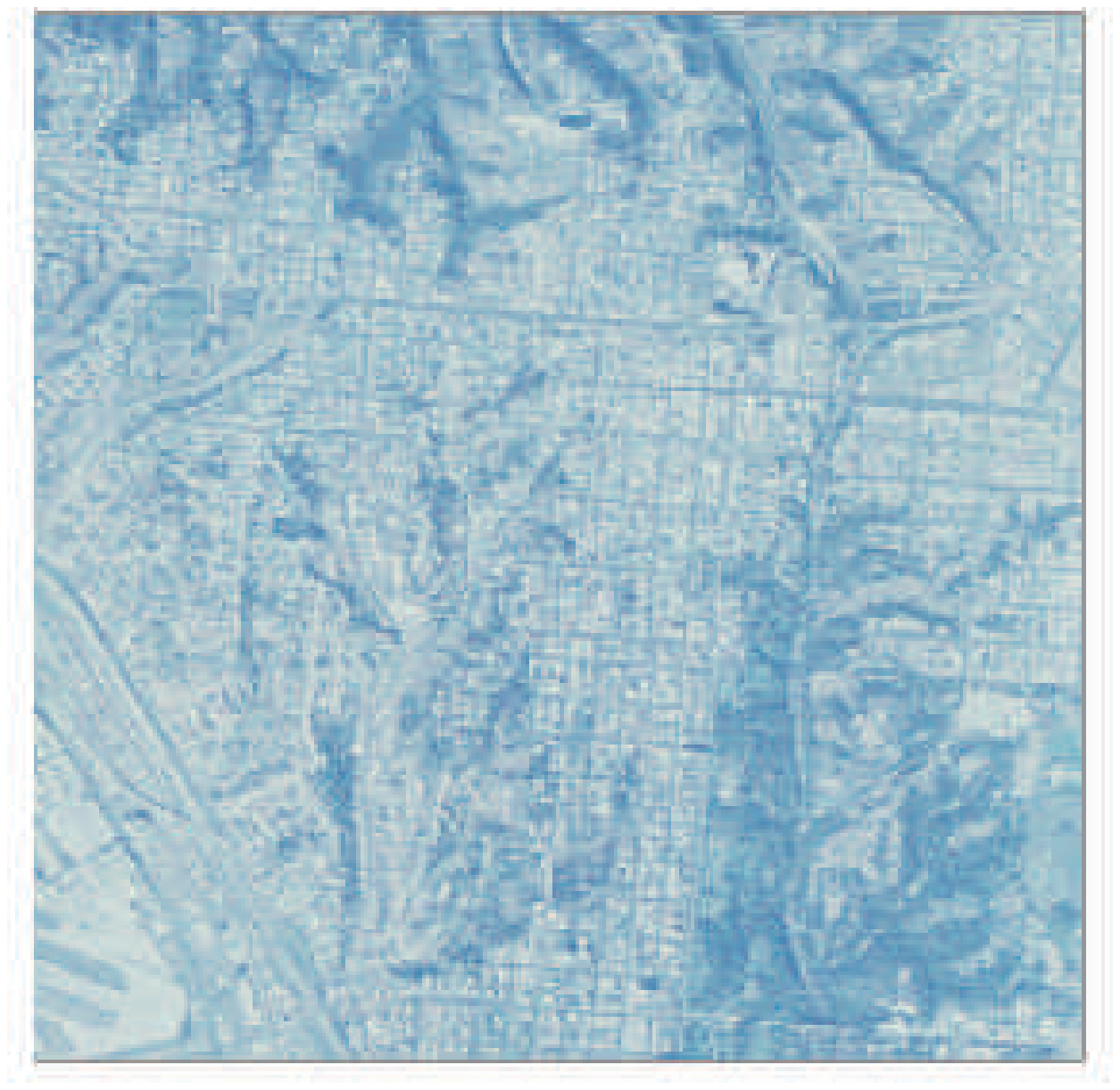}
        \includegraphics[width=0.48\columnwidth,clip=true,trim=120
        45 110 40]{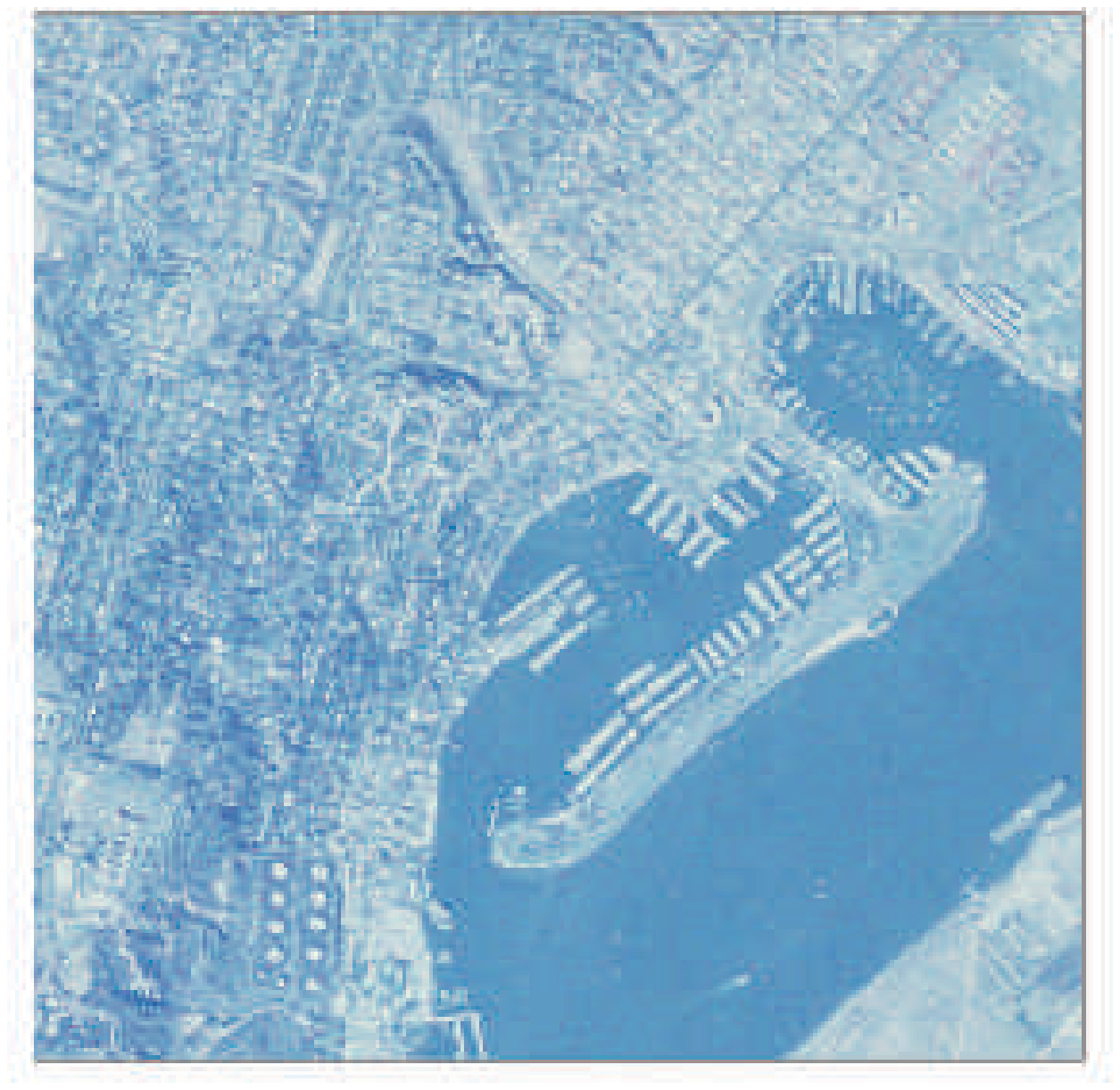}
        \\
        (b) Aerials
    \end{minipage}
    \begin{minipage}{0.32\textwidth}
        \centering
        \includegraphics[width=0.48\columnwidth,clip=true,trim=120
        45 110 40]{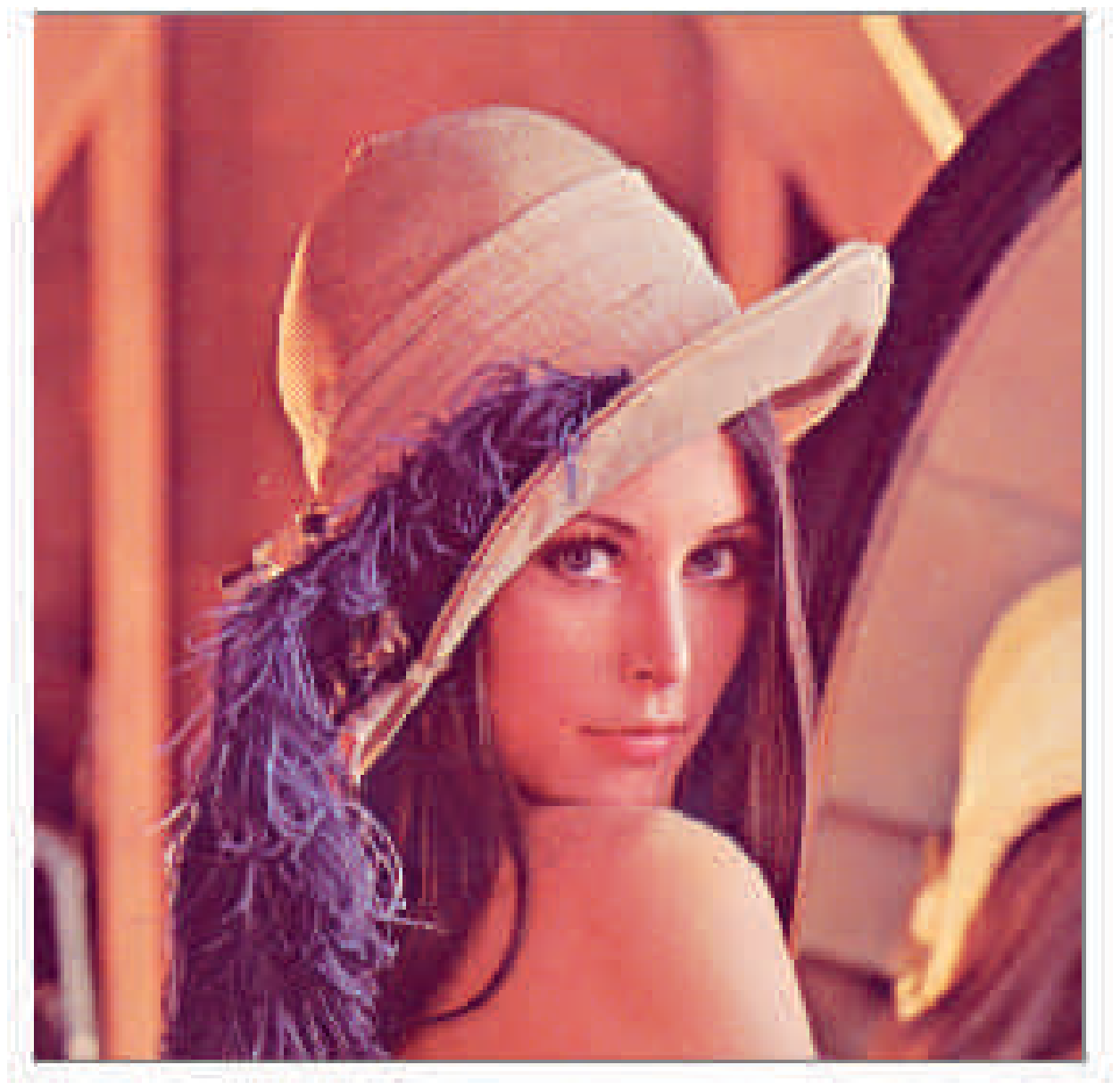}
        \includegraphics[width=0.48\columnwidth,clip=true,trim=120
        45 110 40]{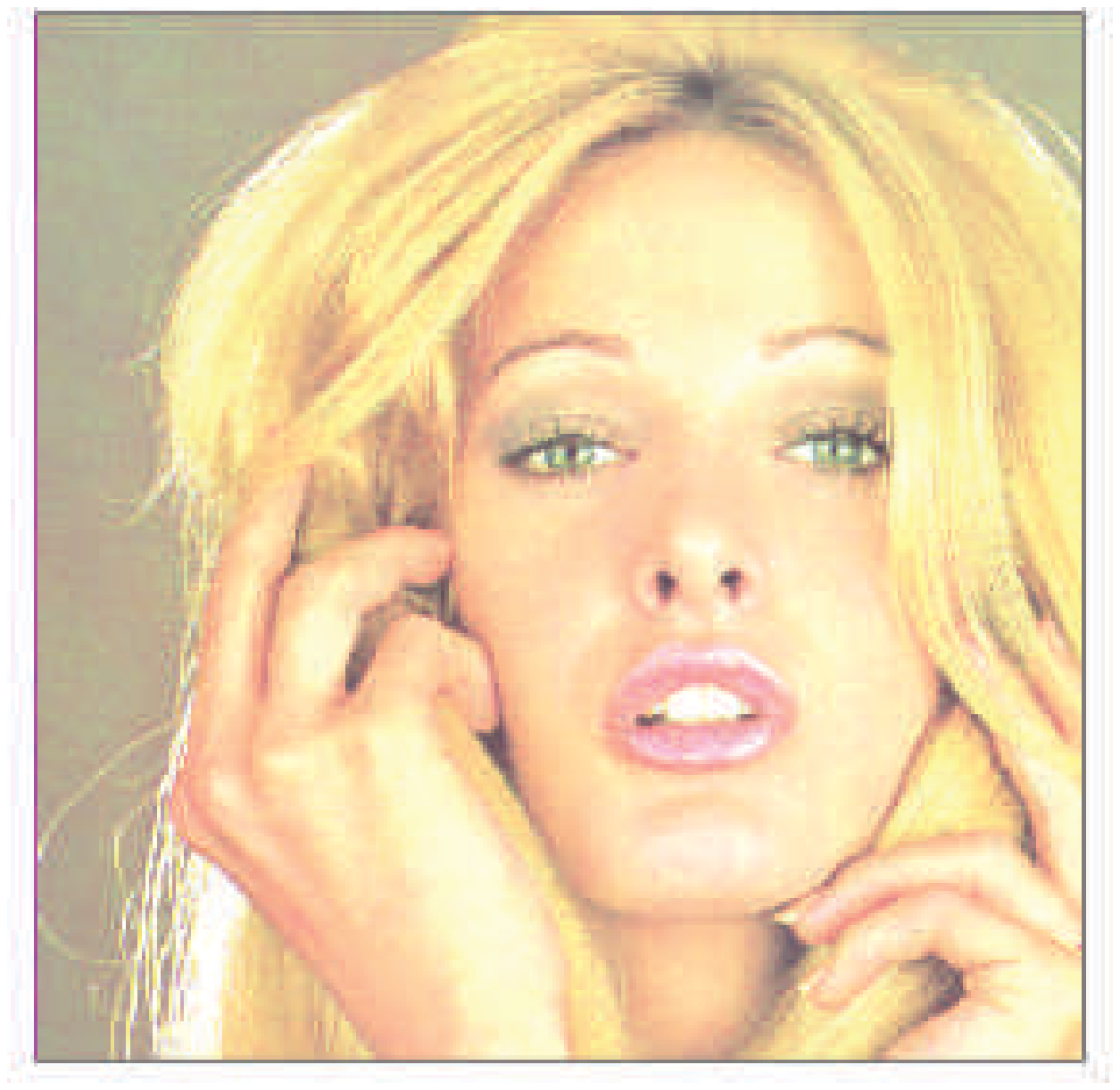}
        \\
        (c) Miscellaneous
    \end{minipage}
    \caption{Sample images from the test image sets}
    \label{fig:test_samples}
    \vskip -0.2in
\end{figure}

We used three sets of images, \textit{textures},
\textit{aerials} and \textit{miscellaneous}, from the
USC-SIPI Image
Database\footnote{http://sipi.usc.edu/database/} as test
images. Tab.~\ref{tab:datasets} lists the details of the
image sets, and Fig.~\ref{fig:test_samples} presents six sample
images from the test sets. 

These datasets are, in terms of contents and properties of
images, very different from each other. For instance, most
of the images in the texture set have highly repetitive
patterns that are not present in the images in the other two
sets. Most images in the aerials set have both coarse
and fine structures, for example, a lake and a nearby road, at the
same time in a single image. Also, the sizes of the images
vary quite a lot across the test sets and across the images
in each set.

\begin{table}[h]
    \centering
    \begin{tabular}{c || c c c c}
        Set & \# of all images & \# of color images & Min.
        Size & Max. Size \\
        \hline
        \hline
        Textures & 64 & 0 & $512 \times 512$ & $1024 \times
        1024$ \\
        Aerials & 38 & 37 & $512 \times 512$ & $2250 \times
        2250$ \\
        Miscellaneous & 44 & 16 & $256 \times 256$ & $1024 \times
        1024$ \\
    \end{tabular}
    \caption{Descriptions of the test image sets.}
    \label{tab:datasets}
\end{table}

As we are aiming to evaluate the performance of denoising a
very general image, we used a large separate data set of
natural image patches to train the models. We extracted a
set of $100,000$ random image patches of sizes $4 \times 4$,
$8 \times 8$ and $16 \times 16$ from CIFAR-10 dataset
\citep{Krizhevsky2009}. From each image of $50,000$ training
samples of the CIFAR-10 dataset, two patches from randomly
selected locations have been collected.

We tried denoising only grayscale images. When an image was
in an RGB format, we averaged the three channels to make the
image grayscale.


\subsection{Denoising Settings}

We tried three different depth settings for both Boltzmann
machines and denoising autoencoders; a single hidden layer,
two hidden layers and four hidden layers. The sizes of all
hidden layers were set to have the same number of hidden
units, which was the constant factor multiplied by the
number of pixels in an image patch\footnote{We used 5 as
suggested by \citet{Xie2012}.}.  

We denote Boltzmann machines with one, two and four hidden
layers by GRBM, GDBM(2) and GDBM(4), respectively. Denoising
autoencoders are denoted by DAE, DAE(2) and DAE(4),
respectively. For each model structure, 
Each model was trained on image patches of sizes $4 \times
4$, $8 \times 8$ and $16 \times 16$.

The GRBMs were trained using the enhanced gradient
\citep{Cho2011icml} and persistent contrastive divergence
(PCD) \citep{Tieleman2008}. The GDBMs were trained by PCD
after initializing the parameters with a two-stage
pretraining algorithm \citep{Cho2012dlufl}. DAEs were
trained by a stochastic backpropagation algorithm, and when
there were more than one hidden layers, we pretrained each
layer as a single-layer DAE with sparsity target set to
$0.1$ \citep{Vincent2010,Lee2007}.

The details on training procedures are described in
Appendix~\ref{sec:training_detail}.

One important difference to the recent work by
\citet{Xie2012} and \citet{Burger2012} is that the denoising
task we consider in this paper is completely \textit{blind}.
No prior knowledge about target images and the type or level
of noise was assumed when training the deep neural networks.
In other words, no separate training was done for different
types or levels of noise injected to the test images. Unlike
this, \citet{Xie2012}, for instance, trained a DAE
specifically for each noise level by changing $\eta(\cdot)$
accordingly.  Furthermore, the Boltzmann machines that we
propose here for image denoising, do not require any prior
knowledge about the level or type of noise.

Two types of noise have been tested; white Gaussian and
salt-and-pepper. White Gaussian noise simply adds zero-mean
normal random value with a predefined variance to each image
pixel, while salt-and-pepper noise sets a randomly chosen
subset of pixels to either black or white. Furthermore,
three different noise levels (0.1, 0.2 and 0.4) were tested.
In the case of white Gaussian noise, they were used as
standard deviations, and in the case of salt-and-pepper
noise, they were used as a noise probability.

After noise was injected, each image was preprocessed by
pixel-wise adaptive Wiener filtering \citep[see,
e.g.,][]{Sonka2007}, following the approach of
\citet{Hyvarinen1999a}.  The width and height of the pixel
neighborhood were chosen to be small enough ($3 \times 3$)
so that it will not remove too much detail from the input
image.  

Denoising performance was measured mainly with peak
signal-to-noise ratio (PSNR) computed by
$
-10 \log_{10} (\epsilon^2),
$
where $\epsilon^2$ is a mean squared error between an
original clean image and the denoised one. 

\subsection{Results and Analysis}

\begin{figure}[t]
    \centering
    \psfrag{DAE}[Bl][Bl][0.7][0]{DAE}
    \psfrag{sDAE2}[Bl][Bl][0.7][0]{DAE(2)}
    \psfrag{sDAE4}[Bl][Bl][0.7][0]{DAE(4)}
    \psfrag{WienerFilt}[Bl][Bl][0.7][0]{Wiener Filt}
    \psfrag{GRBM}[Bc][Bc][0.7][0]{GRBM}
    \psfrag{GDBM2}[Bl][Bl][0.7][0]{GDBM(2)}
    \psfrag{GDBM4}[Bl][Bl][0.7][0]{GDBM(4)}
    \psfrag{GRBM}[Bl][Bl][0.7][0]{GRBM}

    \psfrag{PSNR}[Bc][Bc][0.8][0]{PSNR}
    \psfrag{Noise Level}[Tc][Tc][0.8][0]{Noise Level}
    \psfrag{0.1}[Tc][Tc][0.8][0]{$0.1$}
    \psfrag{0.2}[Tc][Tc][0.8][0]{$0.2$}
    \psfrag{0.3}[Tc][Tc][0.8][0]{$0.3$}

    \begin{minipage}{0.1\textwidth}
        \centering
        $\phantom{\text{\textit{Aerials}}}$
    \end{minipage}
    \begin{minipage}{0.43\textwidth}
        \centering
        White noise
    \end{minipage}
    \begin{minipage}{0.43\textwidth}
        \centering
        Salt-and-pepper noise
    \end{minipage}
    \begin{minipage}{0.1\textwidth}
        \centering
        \textit{Aerials}
    \end{minipage}
    \begin{minipage}{0.43\textwidth}
        \centering
        \includegraphics[width=1\columnwidth]{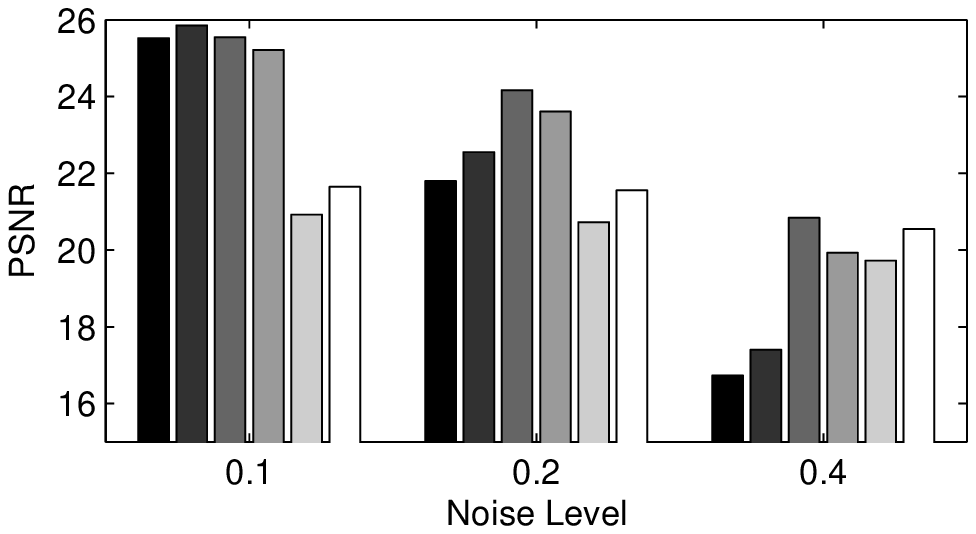}
    \end{minipage}
    \begin{minipage}{0.43\textwidth}
        \centering
        \includegraphics[width=1\columnwidth]{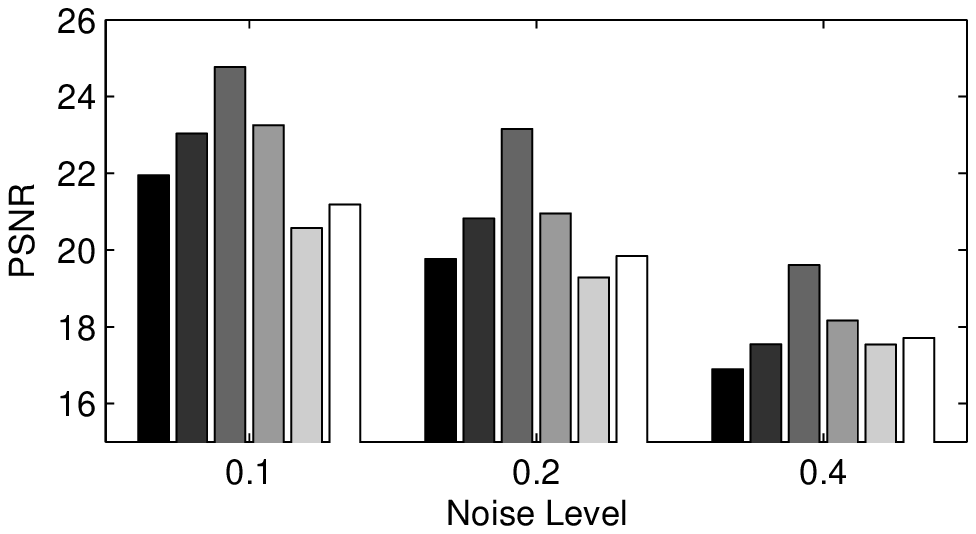}
    \end{minipage}
    \begin{minipage}{0.1\textwidth}
        \centering
        \textit{Textures}
    \end{minipage}
    \begin{minipage}{0.43\textwidth}
        \centering
        \includegraphics[width=1\columnwidth]{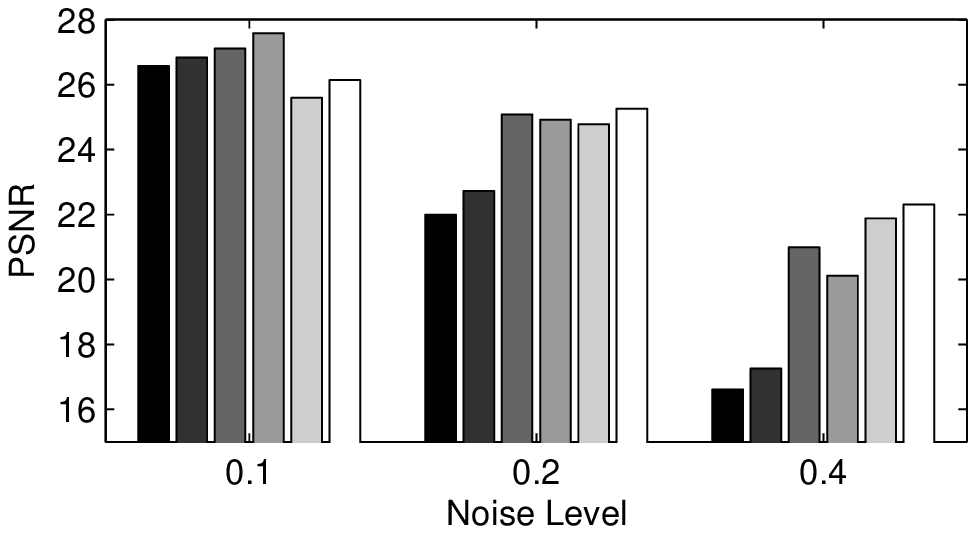}
    \end{minipage}
    \begin{minipage}{0.43\textwidth}
        \centering
        \includegraphics[width=1\columnwidth]{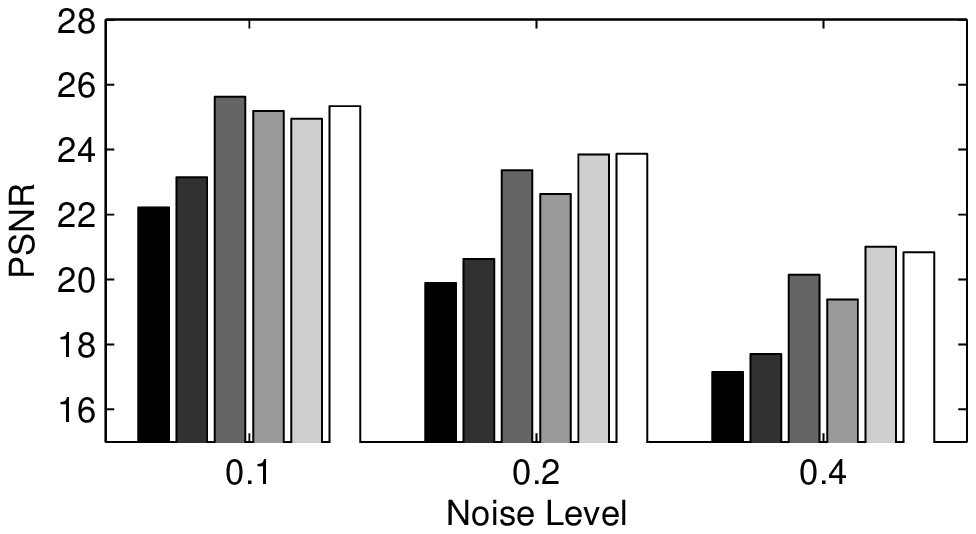}
    \end{minipage}
    \begin{minipage}{0.1\textwidth}
        \centering
        \textit{Misc.}
    \end{minipage}
    \begin{minipage}{0.43\textwidth}
        \centering
        \includegraphics[width=1\columnwidth]{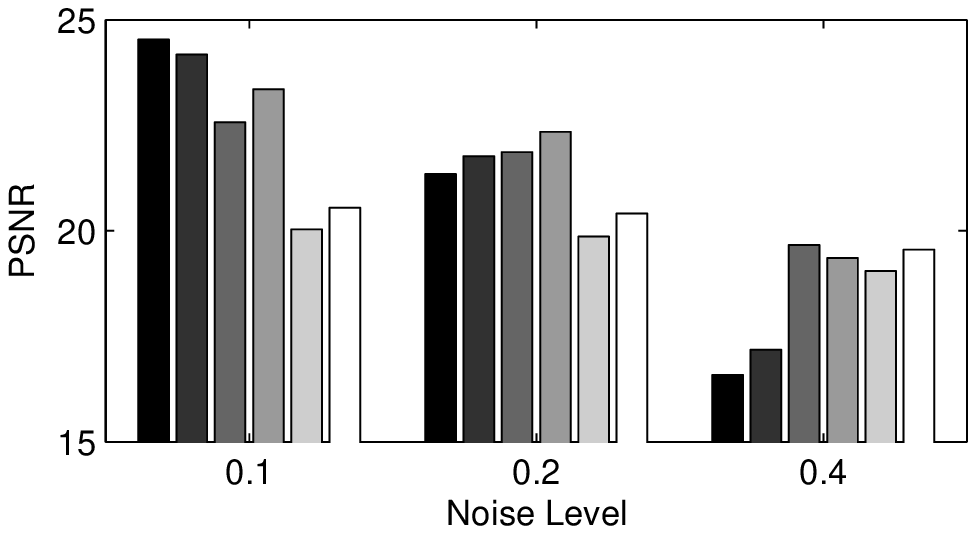}
    \end{minipage}
    \begin{minipage}{0.43\textwidth}
        \centering
        \includegraphics[width=1\columnwidth]{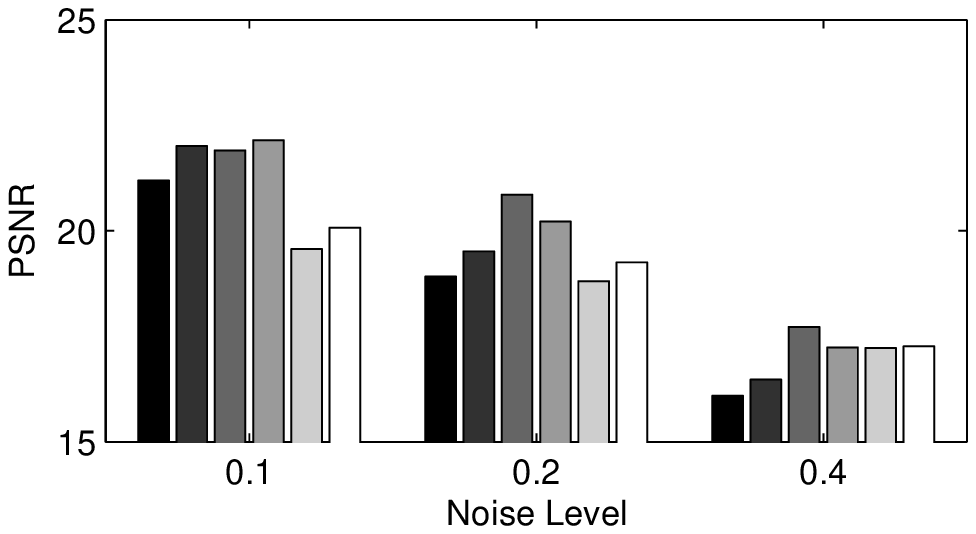}
    \end{minipage}
    \begin{minipage}{0.2\textwidth}
        ~
    \end{minipage}
    \begin{minipage}{0.76\textwidth}
        {
        \centering
        \includegraphics[width=1\columnwidth,clip=true,trim=0
        10 50 0]{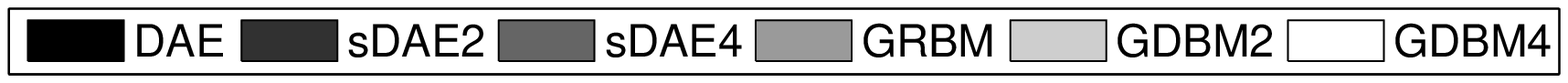}
        }
    \end{minipage}

    \vspace{-6mm}
    \caption{PSNR of grayscale images corrupted by different
    types and levels of noise. The median PSNRs over the
    images in each set together.}
    \label{fig:psnr_all}
    \vskip -0.2in
\end{figure}

In Fig.~\ref{fig:psnr_all}, the performances of all the
tested models trained on $8 \times 8$ image patches are
presented\footnote{Those trained on patches of different
sizes showed similar trend, and they are omitted in this
paper.  }.  

The most obvious observation is that the deep neural
networks, including both DAEs and BMs, did not show 
improvement over their shallow counterparts in the low-noise
regime (0.1). However, the deeper models significantly
outperformed their corresponding shallow models as the level
of injected noise grew. In other words, the power of the
deep models became more evident as the injected level of
noise grew.

This is supported further by Tab.~\ref{tab:psnr} which shows
the performance of the models in the high noise regime
(0.4). In all cases, the deeper models, such as DAE(4),
GDBM(2) and GDBM(4), were the best performing models. 

Another notable phenomenon is that the GDBMs tend to lag
behind the DAEs, and even the GRBM, in the low noise regime,
except for the textures set. A possible explanation for this
rather poor performance of the GDBMs in the low noise regime
is that the approximate inference of the posterior
distribution, used in this experiment, might not have been
good enough.  For instance, more mean-field iterations might
have improved the overall performance while dramatically
increasing the computational time, which would not allow
GDBMs to be of any practical value.  The GDBMs, however,
outperformed, or performed comparably to, the other models
when the level of injected noise was higher.

It should be noticed that the performance depended on
the type of the test images.  For instance, although the
images in the aerials set corrupted with salt-and-pepper
noise were best denoised by the DAE with four hidden layers,
the GDBMs outperformed the DAE(4) in the case of the
textures set. We emphasize here that the deeper neural
networks showed less performance variance depending on the
type of test images, which suggests better generalization
capability of the deeper neural networks.





\begin{figure}[t]
    \centering
    \begin{minipage}{0.99\textwidth}
        \centering
        \begin{minipage}{0.16\columnwidth}
            \centering
            Original
        \end{minipage}
        \begin{minipage}{0.16\columnwidth}
            \centering
            Noisy
        \end{minipage}
        \begin{minipage}{0.16\columnwidth}
            \centering
            DAE
        \end{minipage}
        \begin{minipage}{0.16\columnwidth}
            \centering
            DAE(4)
        \end{minipage}
        \begin{minipage}{0.16\columnwidth}
            \centering
            GRBM
        \end{minipage}
        \begin{minipage}{0.16\columnwidth}
            \centering
            GDBM(4)
        \end{minipage}
    \end{minipage}
    \begin{minipage}{0.99\textwidth}
        \centering
        \begin{minipage}{0.16\columnwidth}
            \includegraphics[width=1\columnwidth,clip=true,trim=120
            45 110 40]{set3img12_orig.eps.jpg.eps}
        \end{minipage}
        \begin{minipage}{0.16\columnwidth}
            \includegraphics[width=1\columnwidth,clip=true,trim=120
            45 110 40]{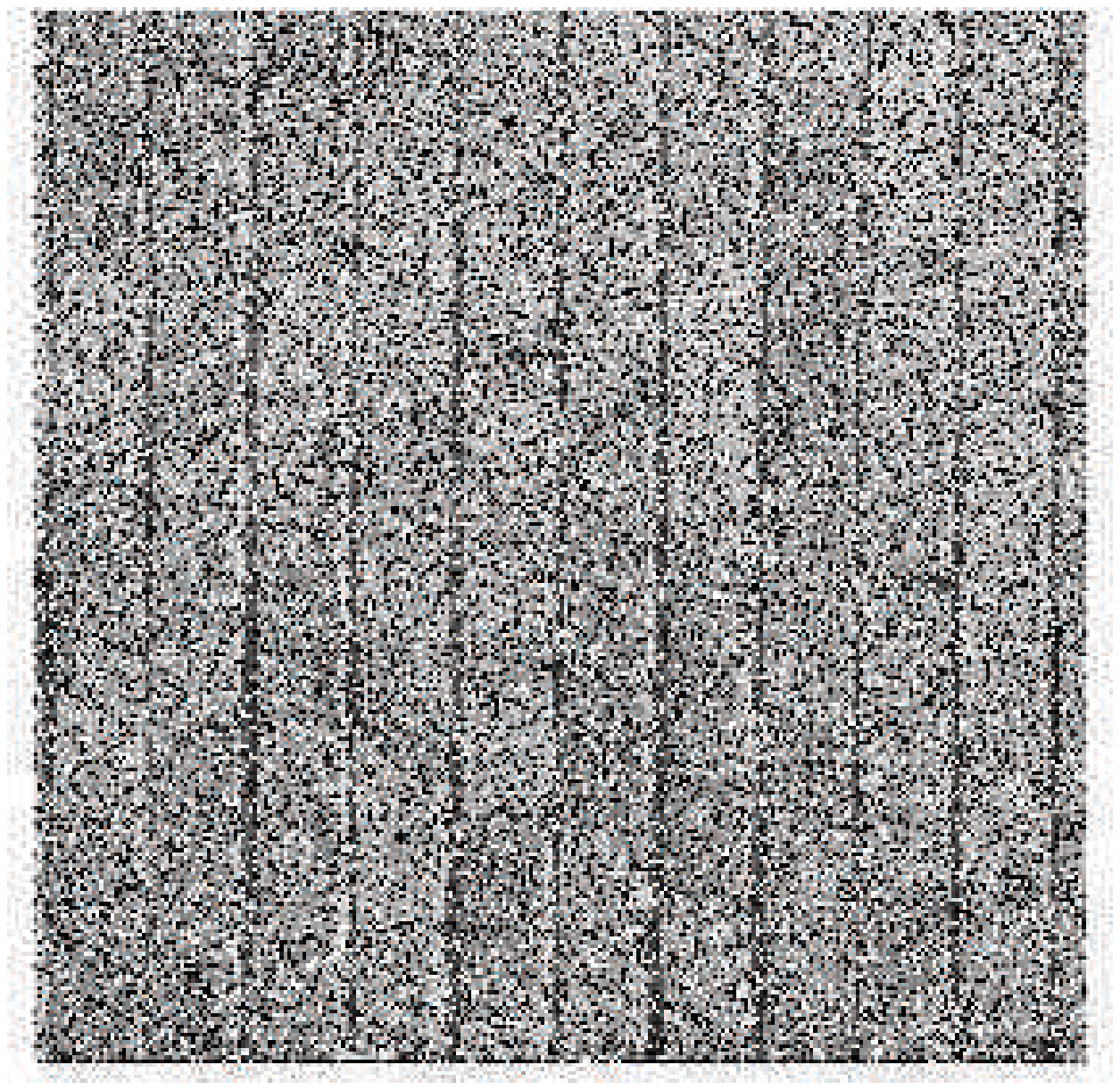}
        \end{minipage}
        \begin{minipage}{0.16\columnwidth}
        \includegraphics[width=1\columnwidth,clip=true,trim=120
        45 110 40]{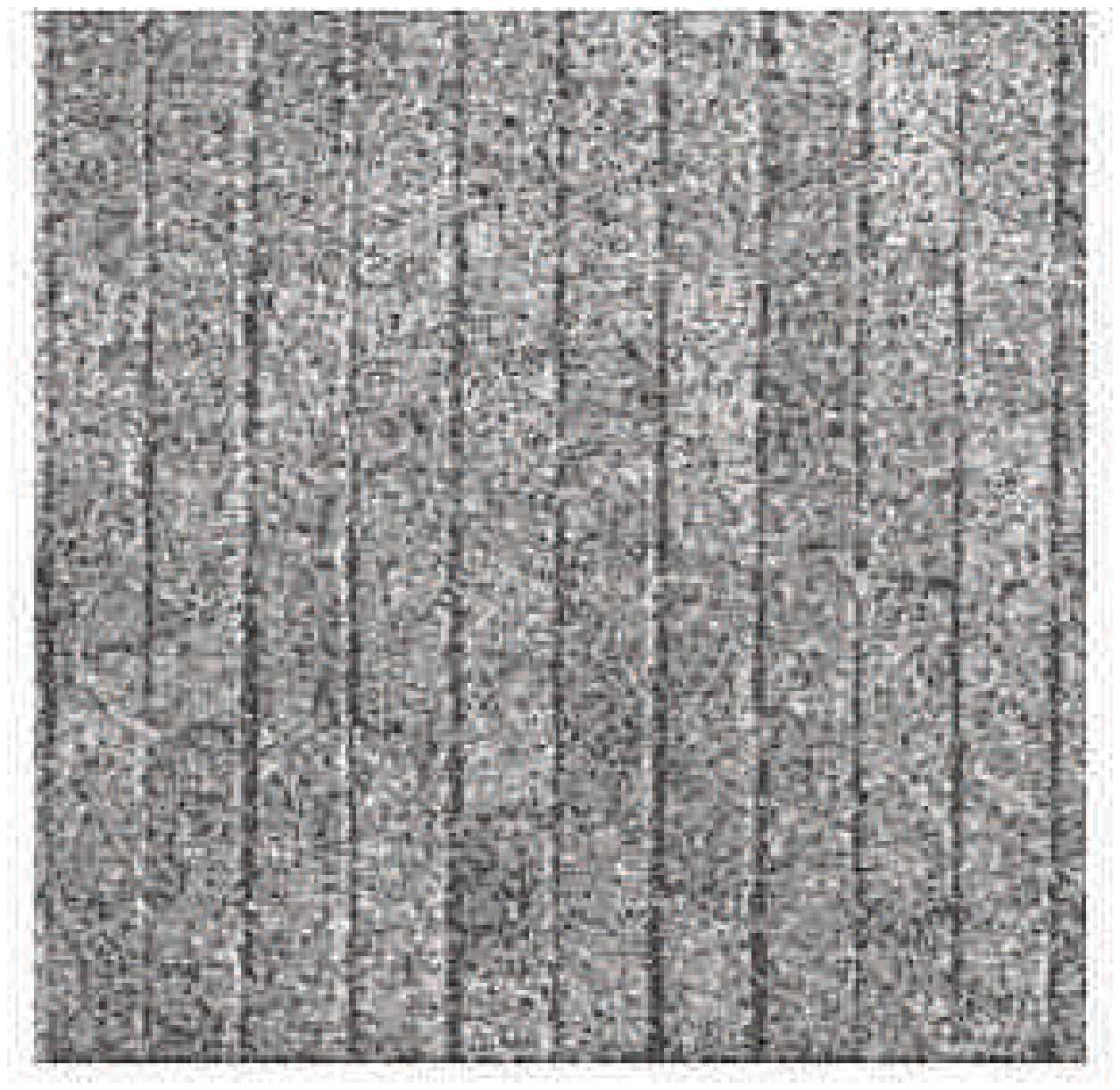}
        \end{minipage}
        \begin{minipage}{0.16\columnwidth}
        \includegraphics[width=1\columnwidth,clip=true,trim=120
        45 110 40]{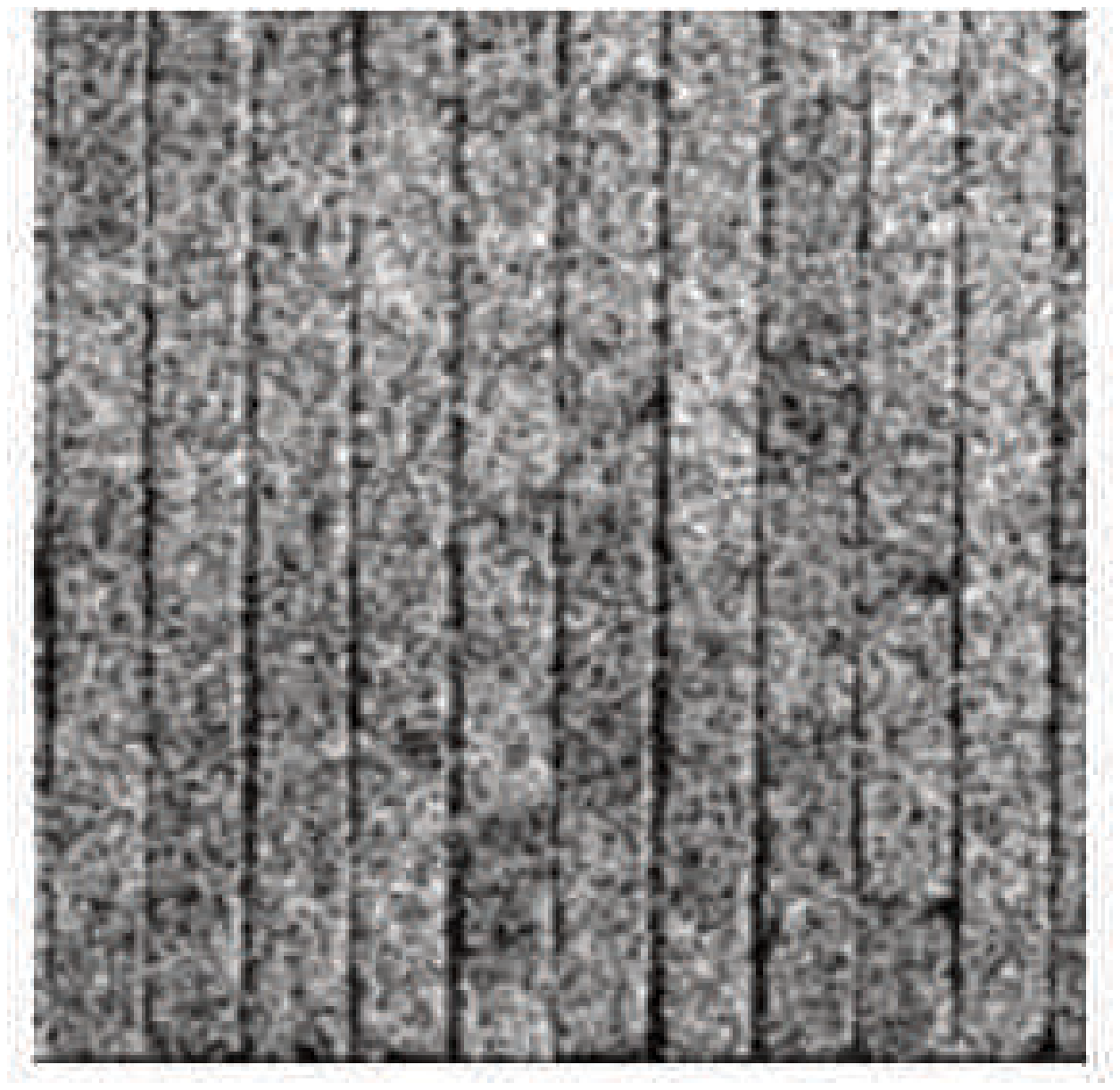}
        \end{minipage}
        \begin{minipage}{0.16\columnwidth}
        \includegraphics[width=1\columnwidth,clip=true,trim=120
        45 110 40]{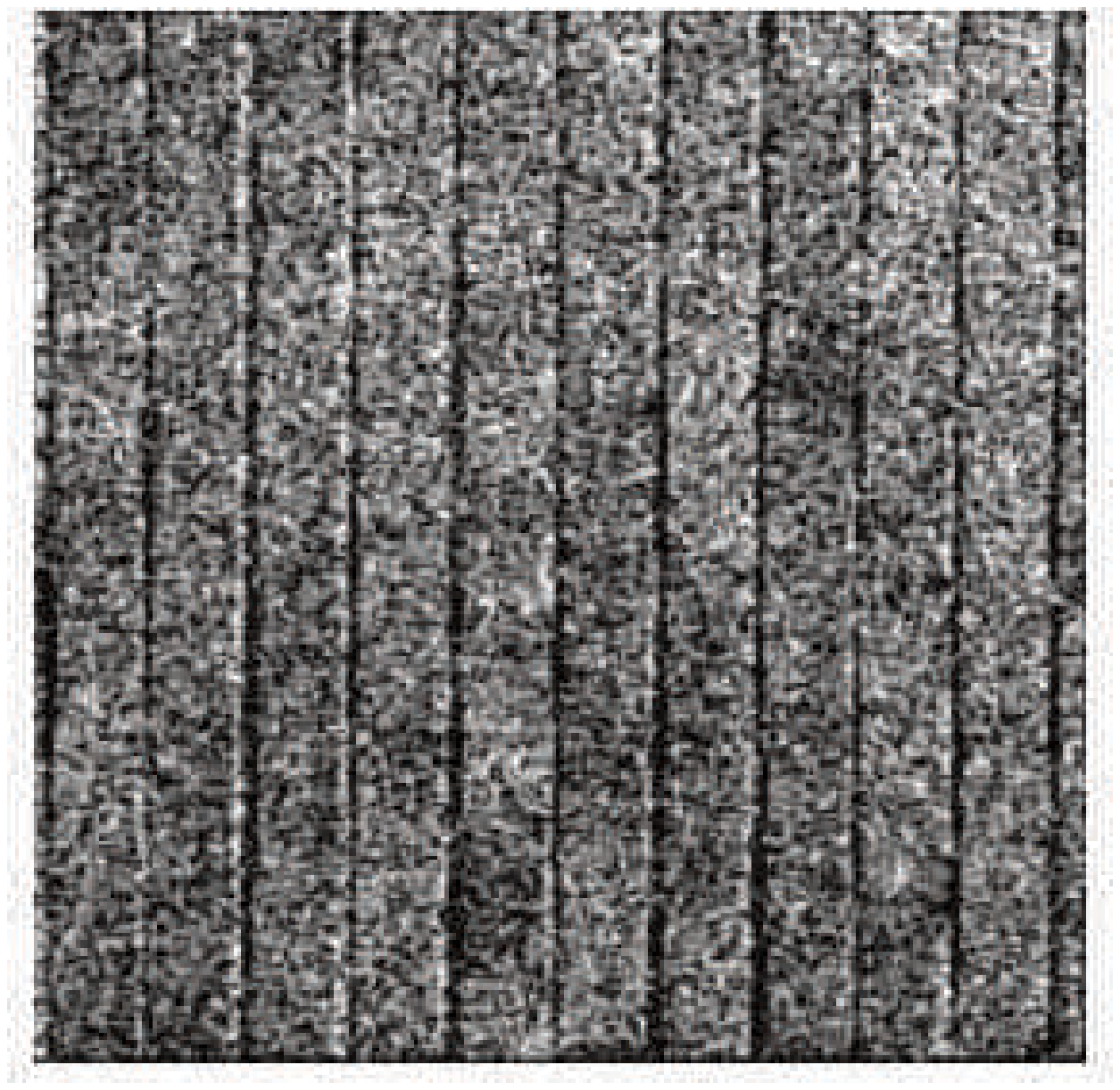}
        \end{minipage}
        \begin{minipage}{0.16\columnwidth}
        \includegraphics[width=1\columnwidth,clip=true,trim=120
        45 110 40]{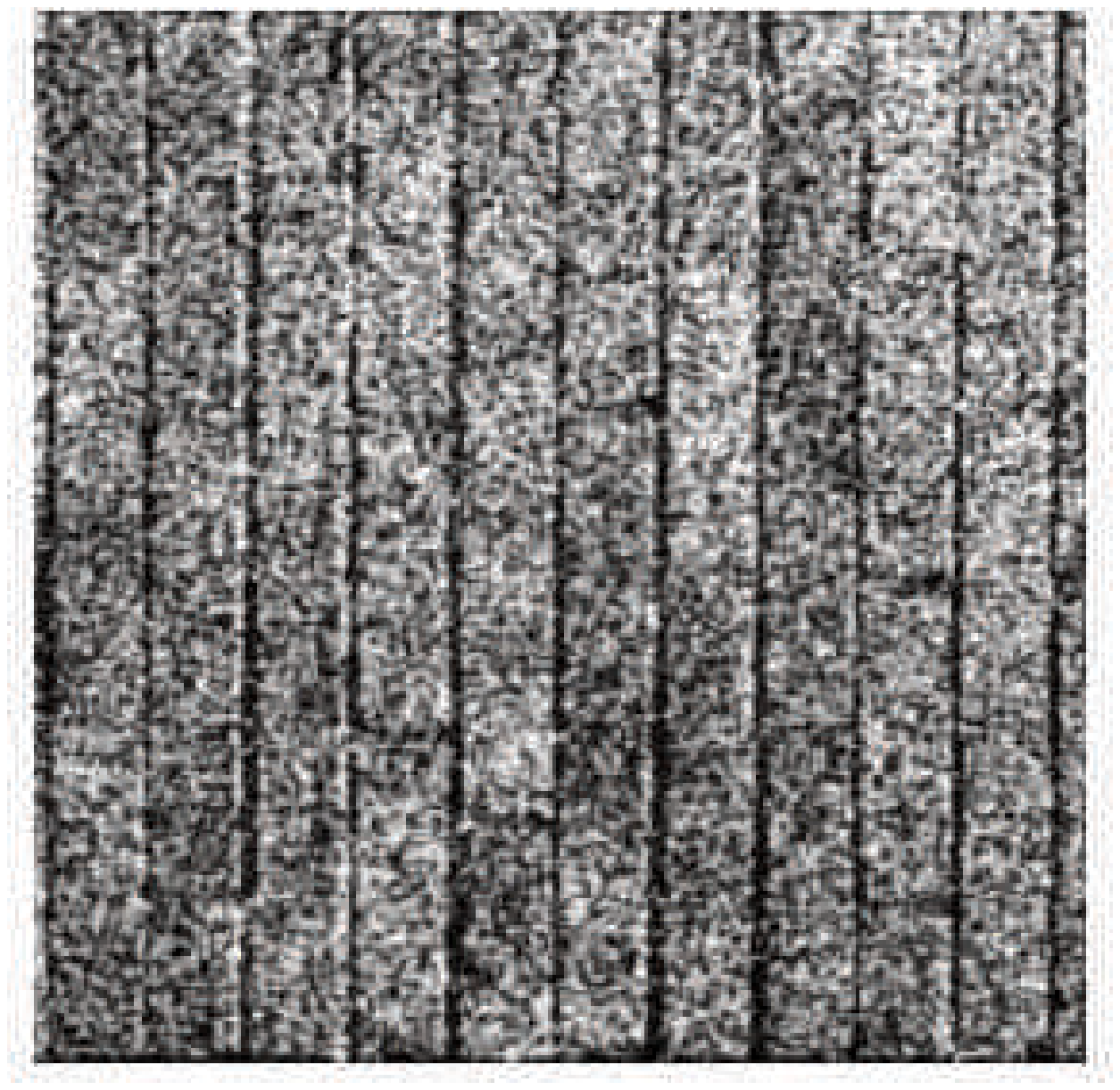}
        \end{minipage}
        \\
        \vspace{0mm}
        {\small
        \begin{minipage}{0.16\columnwidth}
            \centering
            $\phantom{13.91}$
        \end{minipage}
        \begin{minipage}{0.16\columnwidth}
            \centering
            $\phantom{13.91}$
        \end{minipage}
        \begin{minipage}{0.16\columnwidth}
            \centering
            16.79
        \end{minipage}
        \begin{minipage}{0.16\columnwidth}
            \centering
            \textbf{18.96}
        \end{minipage}
        \begin{minipage}{0.16\columnwidth}
            \centering
            18.10
        \end{minipage}
        \begin{minipage}{0.16\columnwidth}
            \centering
            \textbf{18.96}
        \end{minipage}
        }
        \\
        (a) Textures
    \end{minipage}
    \begin{minipage}{0.99\textwidth}
        \centering
        \begin{minipage}{0.16\columnwidth}
        \includegraphics[width=1\columnwidth,clip=true,trim=120
        45 110 40]{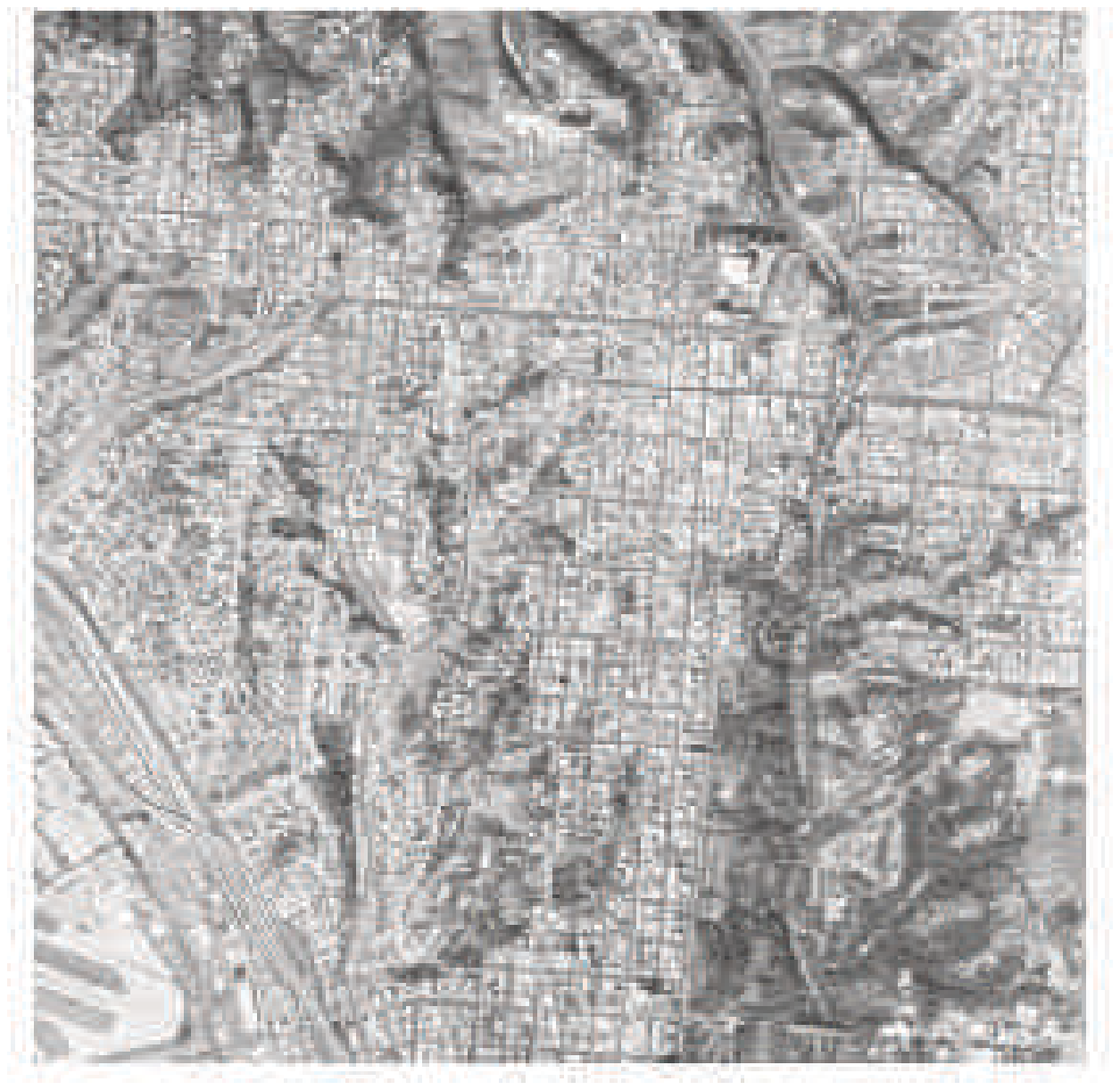}
        \end{minipage}
        \begin{minipage}{0.16\columnwidth}
        \includegraphics[width=1\columnwidth,clip=true,trim=120
        45 110 40]{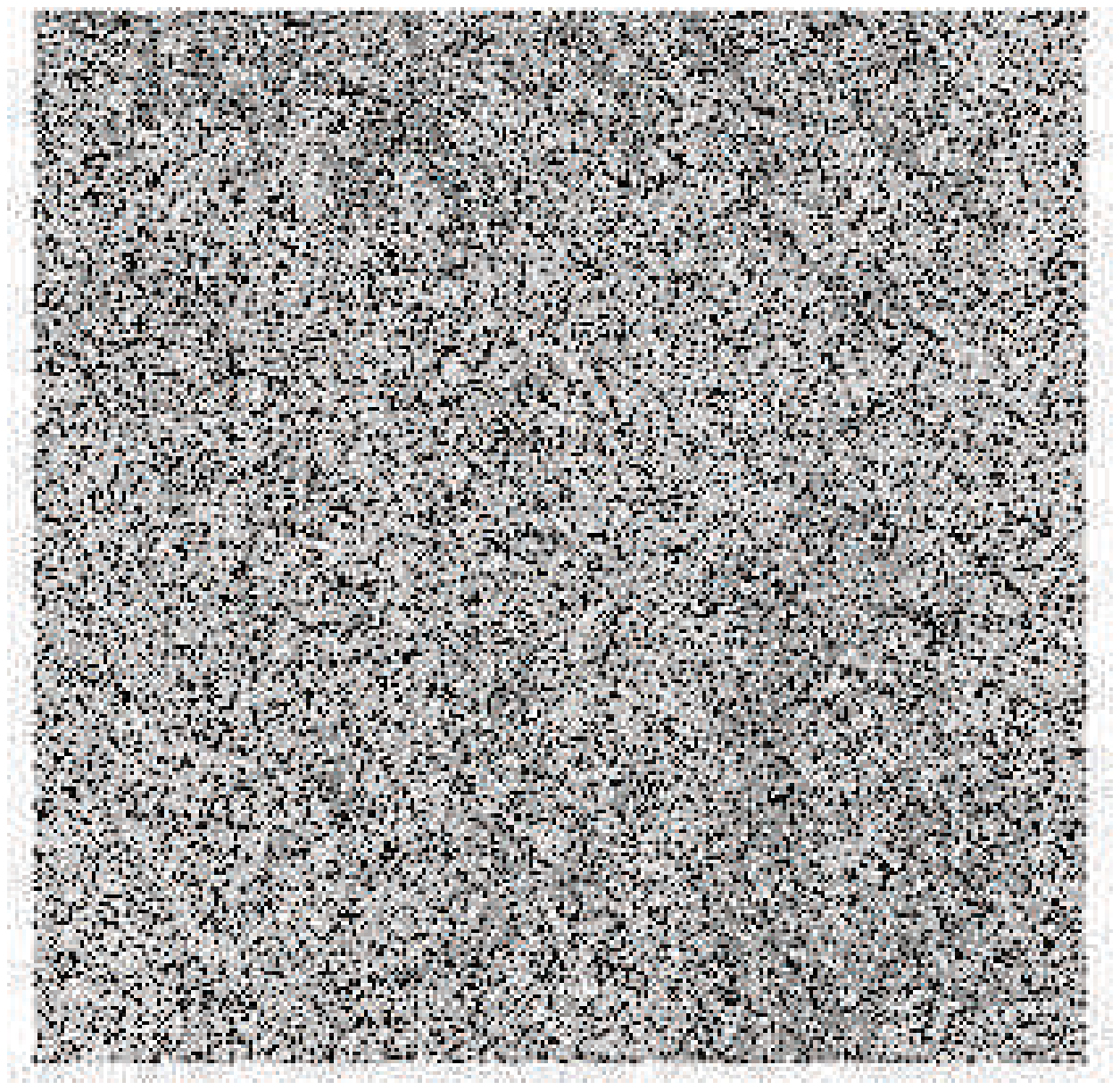}
        \end{minipage}
        \begin{minipage}{0.16\columnwidth}
        \includegraphics[width=1\columnwidth,clip=true,trim=120
        45 110 40]{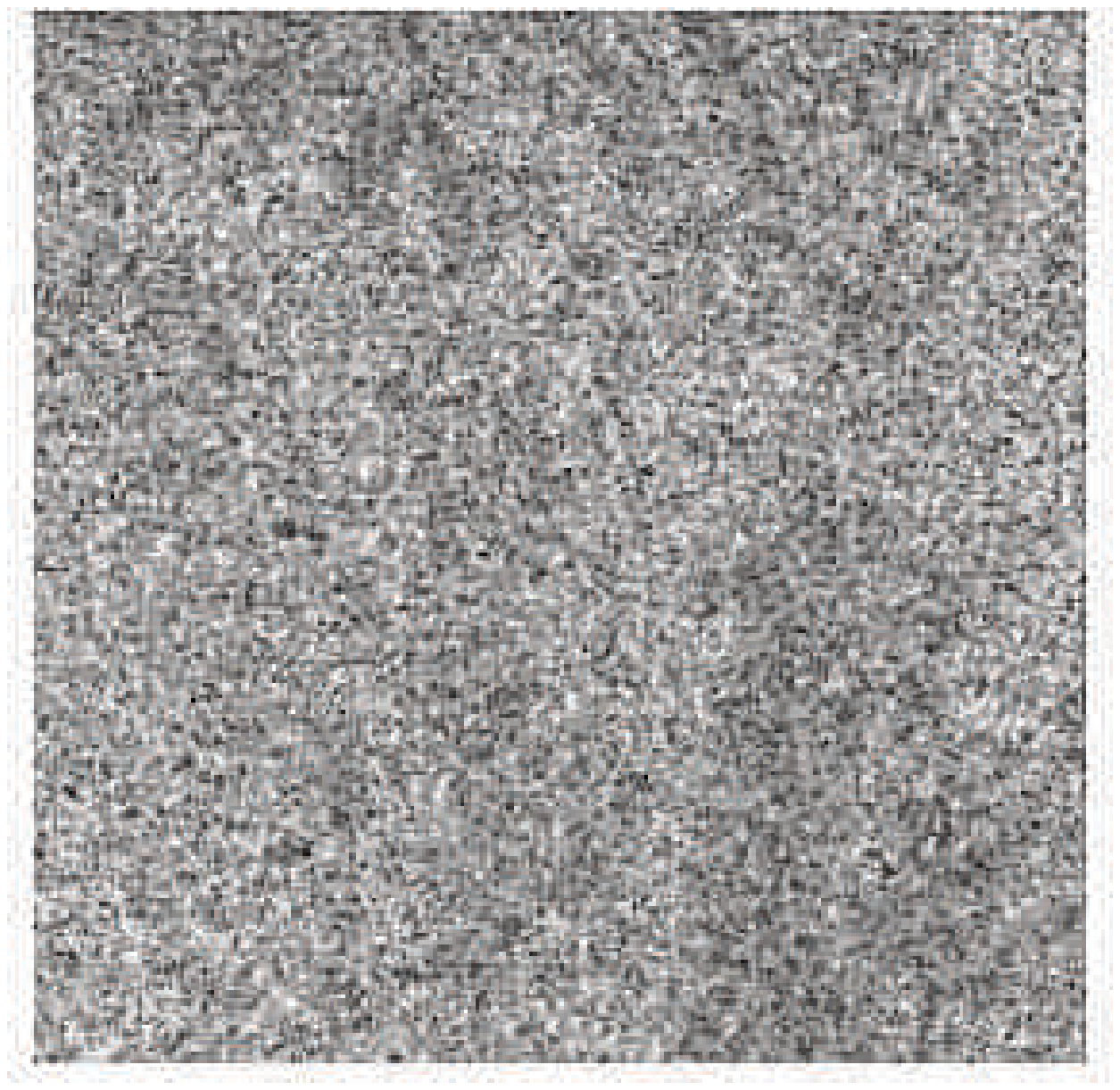}
        \end{minipage}
        \begin{minipage}{0.16\columnwidth}
        \includegraphics[width=1\columnwidth,clip=true,trim=120
        45 110 40]{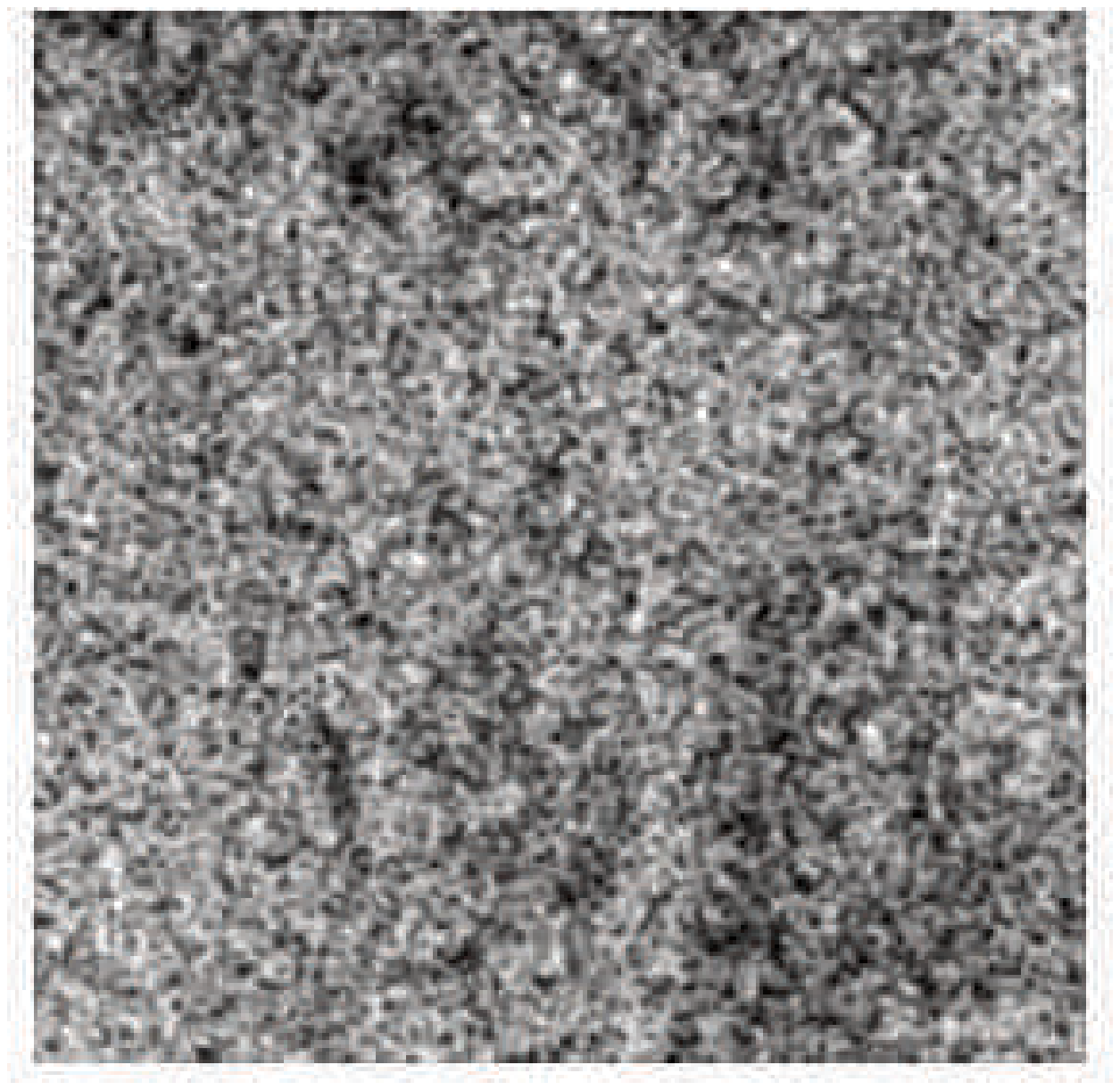}
        \end{minipage}
        \begin{minipage}{0.16\columnwidth}
        \includegraphics[width=1\columnwidth,clip=true,trim=120
        45 110 40]{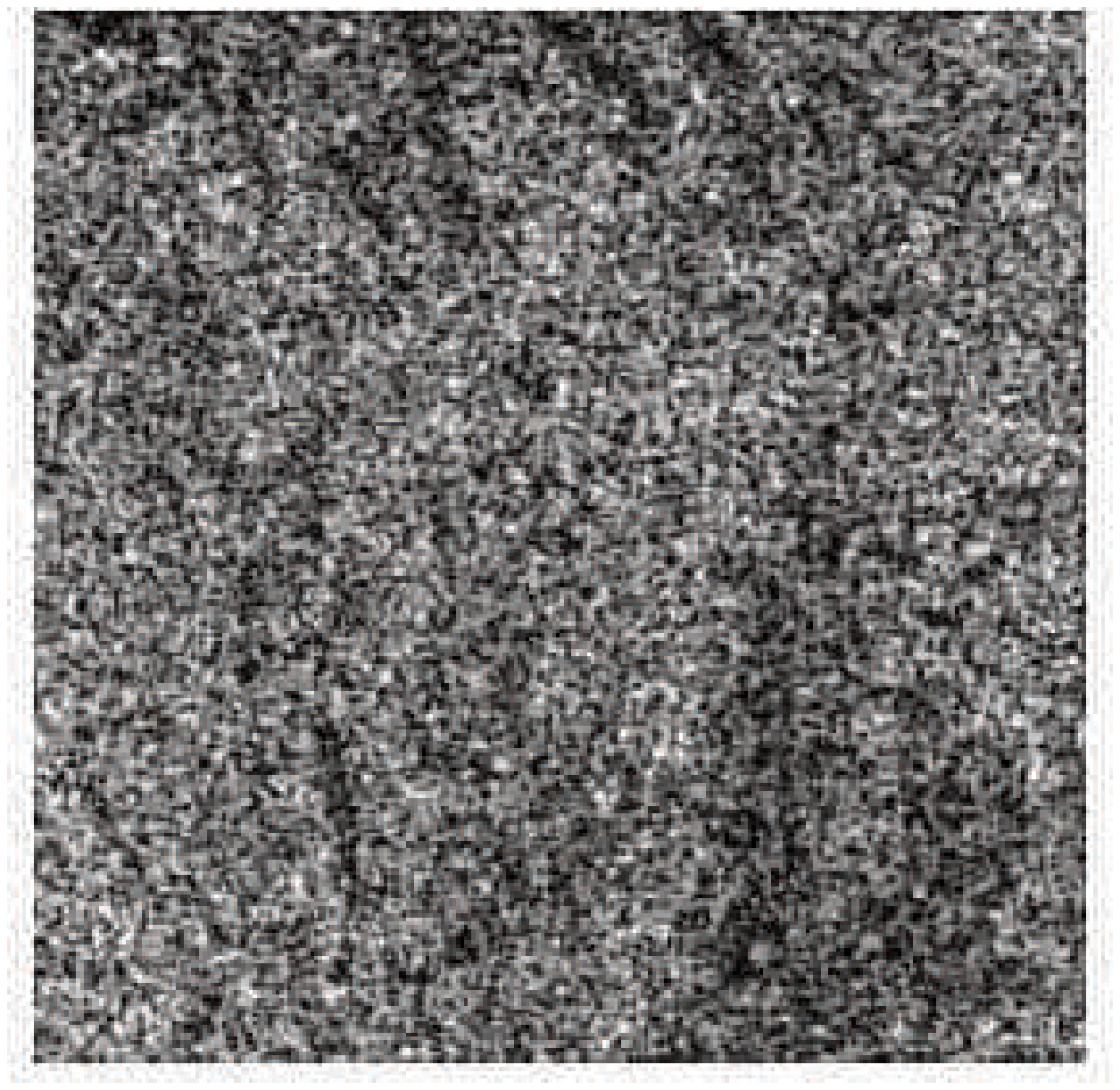}
        \end{minipage}
        \begin{minipage}{0.16\columnwidth}
        \includegraphics[width=1\columnwidth,clip=true,trim=120
        45 110 40]{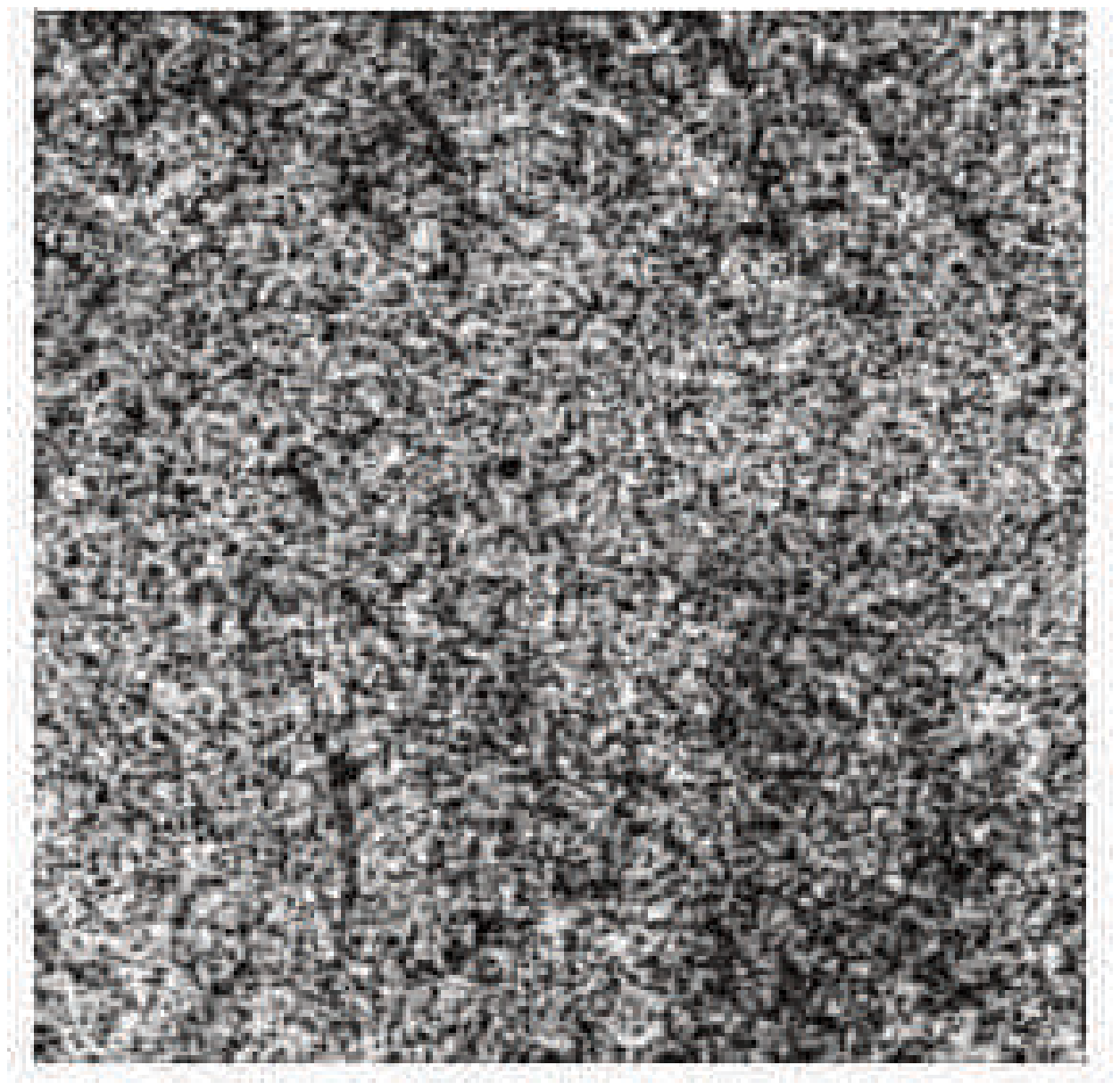}
        \end{minipage}
        \\
        \vspace{0mm}
        {\small
        \begin{minipage}{0.16\columnwidth}
            \centering
            $\phantom{13.91}$
        \end{minipage}
        \begin{minipage}{0.16\columnwidth}
            \centering
            $\phantom{13.91}$
        \end{minipage}
        \begin{minipage}{0.16\columnwidth}
            \centering
            16.02
        \end{minipage}
        \begin{minipage}{0.16\columnwidth}
            \centering
            18.59
        \end{minipage}
        \begin{minipage}{0.16\columnwidth}
            \centering
            17.75
        \end{minipage}
        \begin{minipage}{0.16\columnwidth}
            \centering
            \textbf{19.37}
        \end{minipage}
        }
        \\
        (b) Aerials
    \end{minipage}
    \begin{minipage}{0.99\textwidth}
        \centering
        \begin{minipage}{0.16\columnwidth}
        \includegraphics[width=1\columnwidth,clip=true,trim=120
        45 110 40]{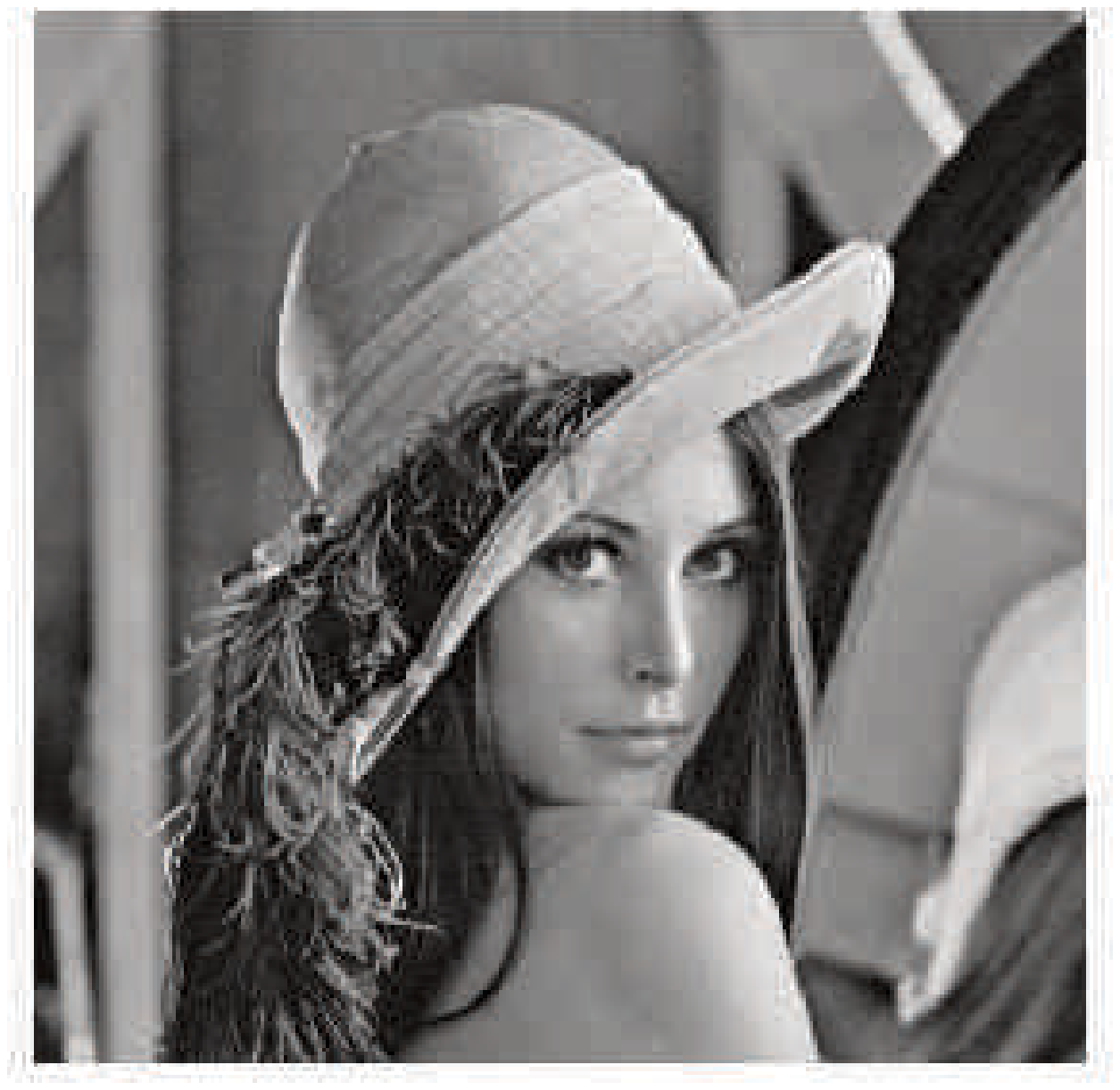}
        \end{minipage}
        \begin{minipage}{0.16\columnwidth}
        \includegraphics[width=1\columnwidth,clip=true,trim=120
        45 110 40]{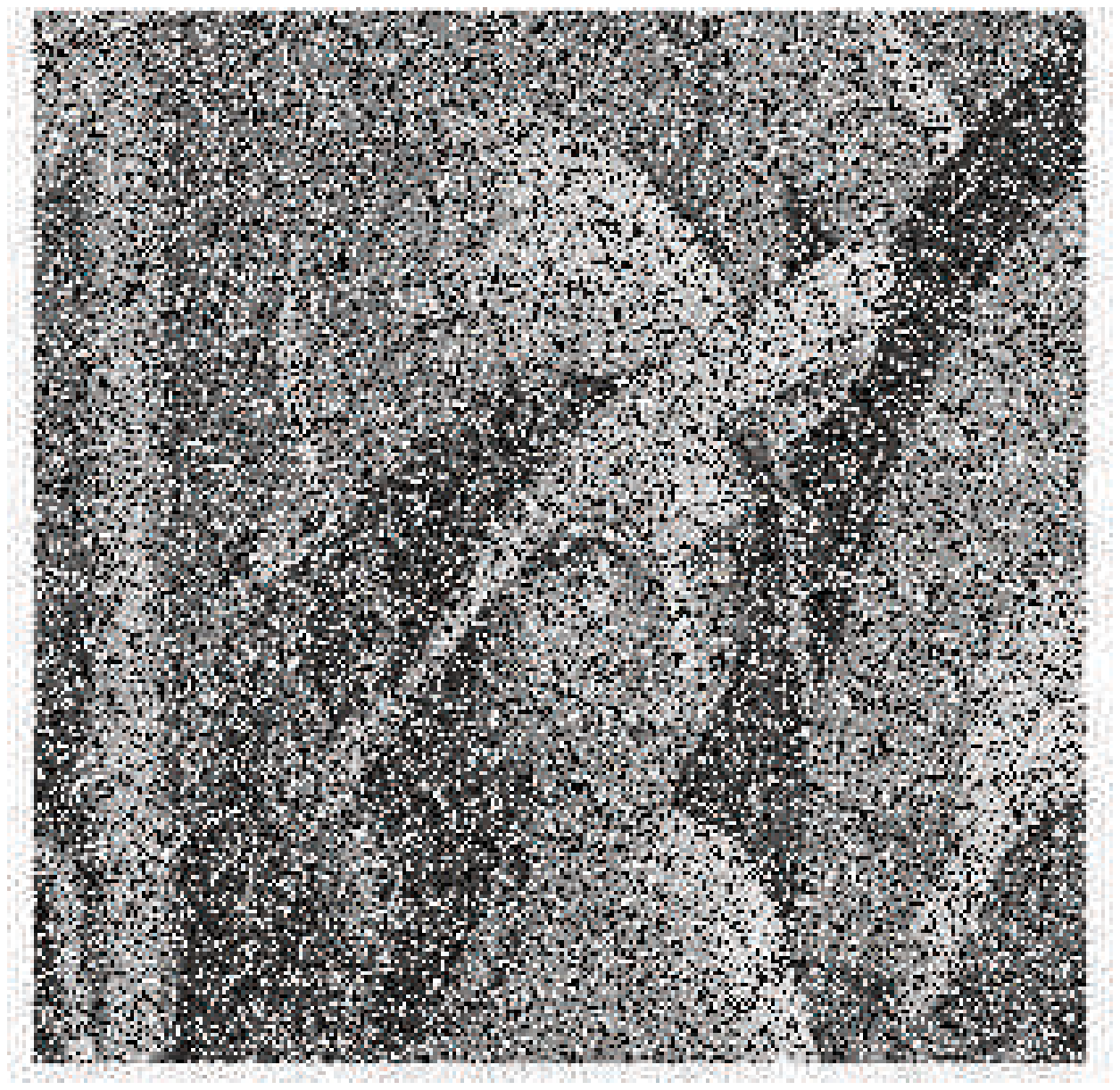}
        \end{minipage}
        \begin{minipage}{0.16\columnwidth}
        \includegraphics[width=1\columnwidth,clip=true,trim=120
        45 110 40]{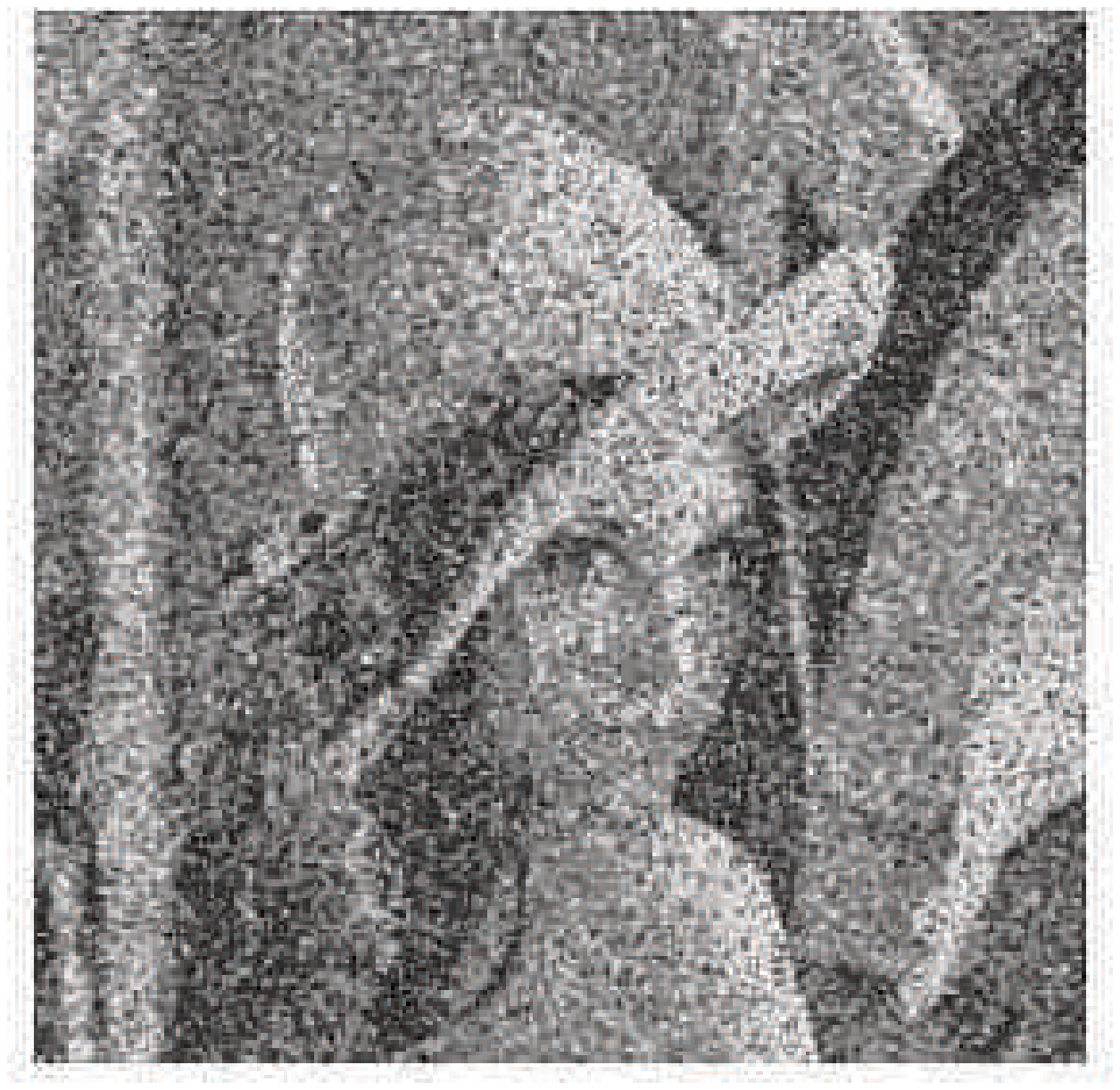}
        \end{minipage}
        \begin{minipage}{0.16\columnwidth}
        \includegraphics[width=1\columnwidth,clip=true,trim=120
        45 110 40]{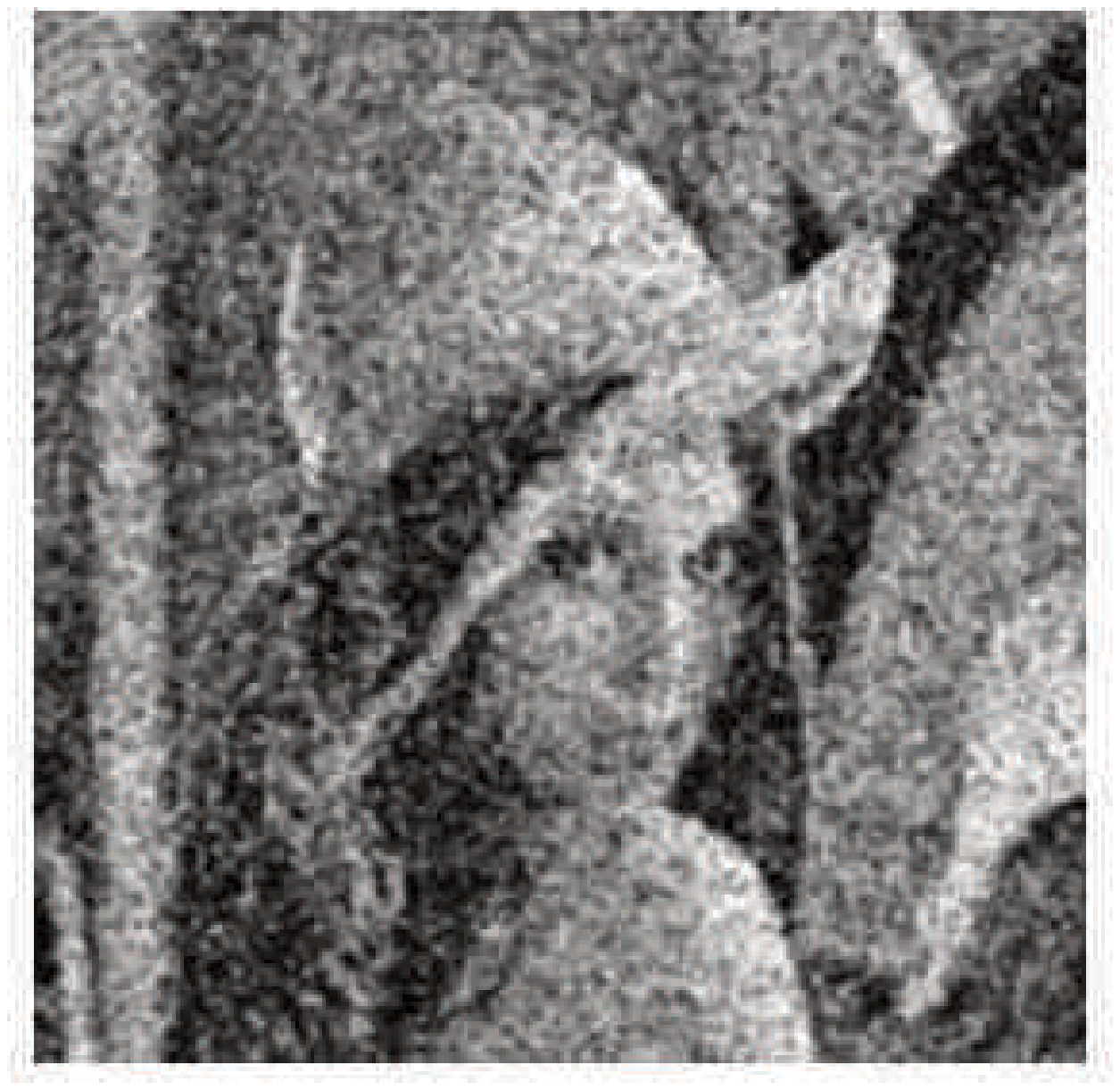}
        \end{minipage}
        \begin{minipage}{0.16\columnwidth}
        \includegraphics[width=1\columnwidth,clip=true,trim=120
        45 110 40]{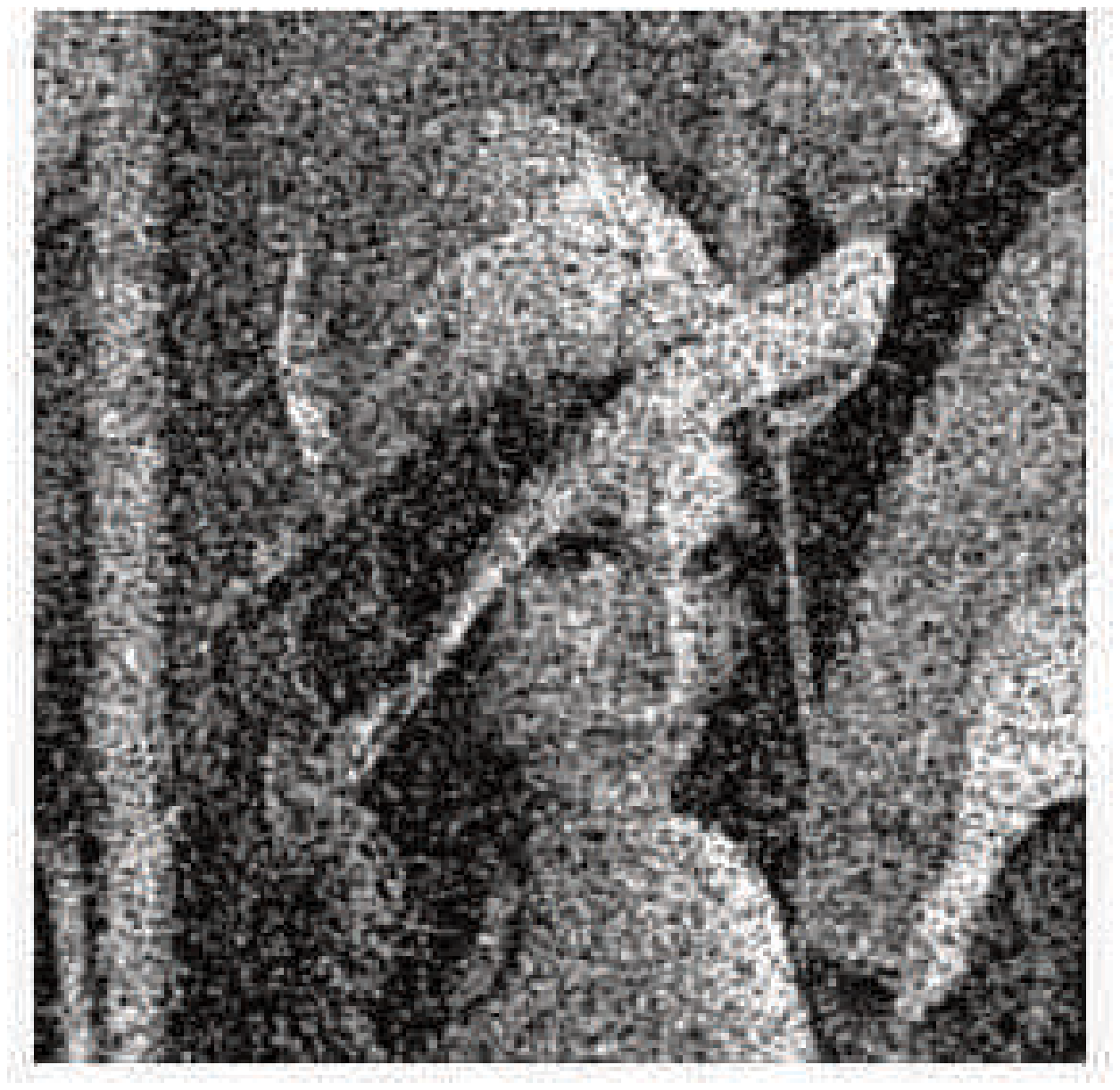}
        \end{minipage}
        \begin{minipage}{0.16\columnwidth}
        \includegraphics[width=1\columnwidth,clip=true,trim=120
        45 110 40]{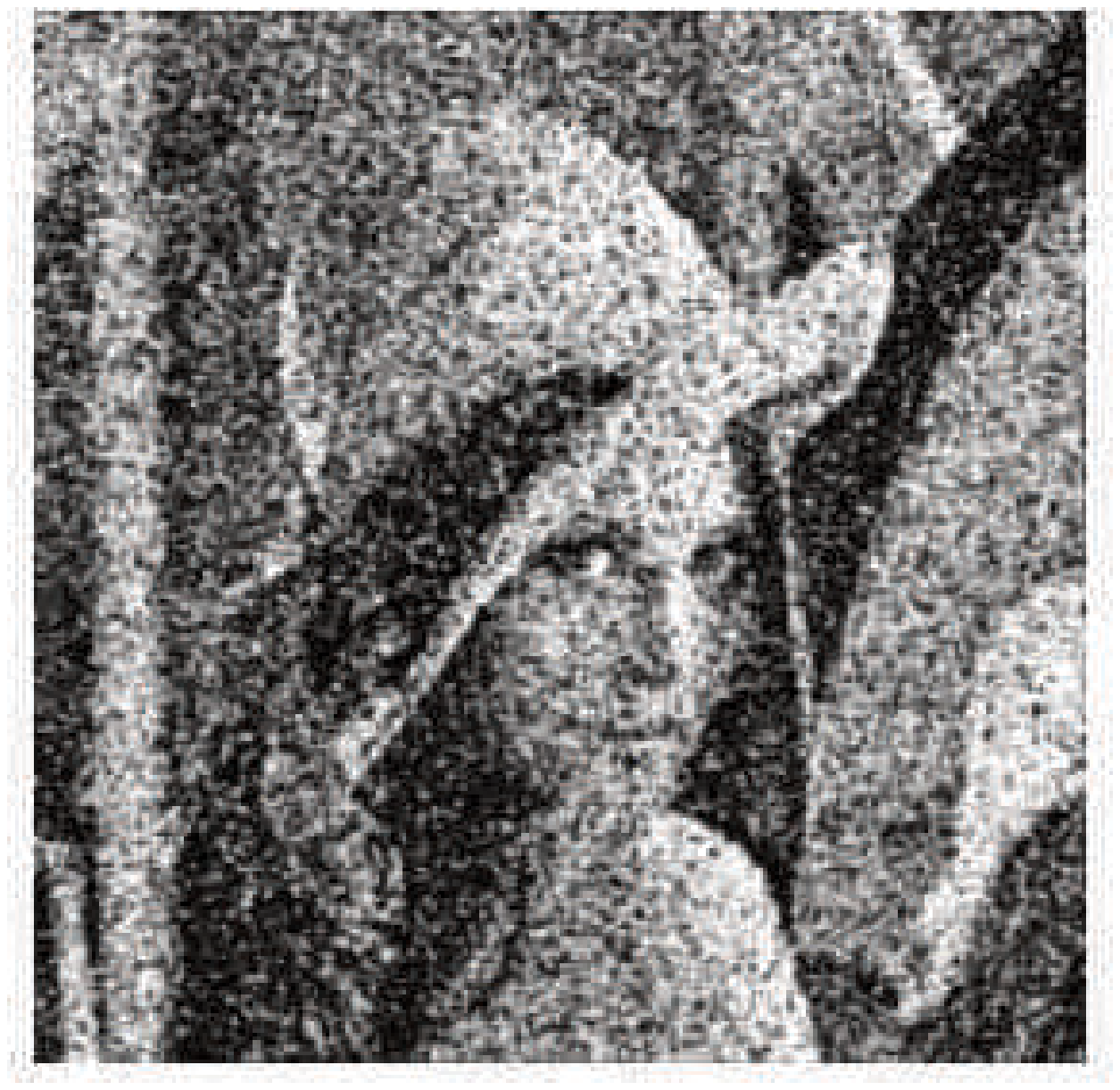}
        \end{minipage}
        \\
        \vspace{0mm}
        {\small
        \begin{minipage}{0.16\columnwidth}
            \centering
            $\phantom{13.91}$
        \end{minipage}
        \begin{minipage}{0.16\columnwidth}
            \centering
            $\phantom{13.91}$
        \end{minipage}
        \begin{minipage}{0.16\columnwidth}
            \centering
            17.59
        \end{minipage}
        \begin{minipage}{0.16\columnwidth}
            \centering
            \textbf{21.34}
        \end{minipage}
        \begin{minipage}{0.16\columnwidth}
            \centering
            19.42
        \end{minipage}
        \begin{minipage}{0.16\columnwidth}
            \centering
            18.98
        \end{minipage}
        }
        \\
        (c) Miscellaneous
    \end{minipage}
    \caption{Images, corrupted by salt-and-pepper noise with
    0.4 noise probability, denoised by various deep neural
    networks trained on $8 \times 8$ image patches. The
    number below each denoised image is the PSNR.}
    \label{fig:denoised}
    \vskip -0.2in
\end{figure}

Visual inspection of the denoised images provides some more
intuition on the performances of the deep neural networks.
In Fig.~\ref{fig:denoised}, the denoised images of a sample
image from each test image set are displayed. It shows that
BMs tend to emphasize the detailed structure of the image,
while DAEs, especially ones with more hidden layers, tend to
capture the global structure. 


Additionally, we tried the same set of experiments using the
models trained on a set of $50,000$ random image patches
extracted from the Berkeley Segmentation Dataset
\citet{Martin2001}. In this case, $100$ patches from
randomly chosen locations from each of $500$ images were
collected to form the training set. We obtained the results
similar to those presented in this paper. The results are
presented in Appendix~\ref{sec:bsd500}.

\begin{table}[ht]
    \centering
    \begin{minipage}{0.48\textwidth}
        \centering
        \begin{tabular}{r || c c c}
            Method & Aerials & Textures & Misc. \\
            \hline
            Wiener &  15.7 \tiny{(0.1)} & 15.5 \tiny{(0.6)} &15.9 \tiny{(0.6)} \\
            DAE &     16.4 \tiny{(0.2)} & 16.2 \tiny{(0.9)} &16.6 \tiny{(0.8)} \\
            DAE(2) &  17.6 \tiny{(0.2)} & 17.1 \tiny{(1.2)} &17.7 \tiny{(1.1)} \\
            DAE(4) &  20.8 \tiny{(0.7)} & \textbf{18.7}
            \tiny{(2.8)} & \textbf{20.2} \tiny{(2.0)} \\
            GRBM &    19.2 \tiny{(0.4)} & 18.0 \tiny{(1.7)} &18.9 \tiny{(1.5)} \\
            GDBM(2) & \textbf{22.3} \tiny{(1.4)} & \textbf{18.7} \tiny{(3.2)} &20.1 \tiny{(2.4)} \\
            GDBM(4) & 22.1 \tiny{(1.1)} & \textbf{18.7}
            \tiny{(3.0)} & \textbf{20.2} \tiny{(2.2)} \\
            \hline
        \end{tabular}
    \end{minipage}
    \begin{minipage}{0.48\textwidth}
        \centering
        \begin{tabular}{c || c c c}
            & Aerials & Textures & Misc. \\
            \hline
            & 16.3 \tiny{(0.5)} &14.9 \tiny{(1.3)} &  15.3 \tiny{(1.5)} \\
            & 17.1 \tiny{(0.6)} &15.7 \tiny{(1.4)} &  16.3 \tiny{(1.7)} \\
            & 18.1 \tiny{(0.7)} &16.4 \tiny{(1.7)} &  17.3 \tiny{(2.0)} \\
            & 20.1 \tiny{(1.1)} &\textbf{17.2} \tiny{(2.8)}
            &  \textbf{19.0} \tiny{(2.7)} \\
            & 18.9 \tiny{(0.9)} &16.6 \tiny{(2.1)} &  17.6 \tiny{(2.2)} \\
            & \textbf{20.3} \tiny{(1.4)} &16.5 \tiny{(3.0)} &  17.5 \tiny{(2.6)} \\
            & \textbf{20.3} \tiny{(1.3)} &16.6 \tiny{(2.9)} &  17.6 \tiny{(2.5)} \\
            \hline
        \end{tabular}
    \end{minipage}
    \\
    \begin{minipage}{0.48\textwidth}
        \centering
        (a) White Gaussian Noise
    \end{minipage}
    \begin{minipage}{0.48\textwidth}
        \centering
        (b) Salt-and-Pepper Noise
    \end{minipage}
    \caption{Performance of the models trained on $4 \times
    4$ image patches when the level of
    injected noise was 0.4. Standard deviations are shown
    inside the parentheses, and the best performing models
    are marked bold.}
    \label{tab:psnr}
\end{table}

\section{Conclusion}

In this paper, we proposed that, in addition to DAEs,
Boltzmann machines, GRBMs and GDBMS, can also be used for
denoising images. Furthermore, we tried to find empirical
evidence supporting the use of deep neural networks in image
denoising tasks.

Our experiments suggest the following conclusions for the
questions raised earlier:

\textbf{Does a model with more hidden layers perform
better?} 

In the case of DAEs, the experiments clearly show that more
hidden layers do improve performance, especially when the
level of noise is high. This does not always apply to BMs,
where we found that the GRBMs outperformed, or performed as
well as, the GDBMs in few cases. Regardlessly, in the high
noise regime, it was always beneficial to have more hidden
layers.

\textbf{How well does a deep model generalize?} 

The deep neural networks were trained on a completely
separate dataset and were applied to three test sets with
very different image properties. It turned out that the
performance depended on each test set, however, with only
small differences. Also, the trend of deeper models
performing better could be observed in almost all cases,
again especially with high level of noise. This suggests
that a well-trained deep neural network can perform
\textit{blind} image denoising, where no prior information
about target, noisy images is available, well.

\textbf{Which family of deep neural networks is more
suitable, BMs or DAEs?}

The DAE with four hidden layers turned out to be the best
performer, in general, beating GDBMs with the same number of
hidden layers. However, when the level of noise was high,
the Boltzmann machines such as GRBM and GDBM(2) were able to
outperform the DAEs, which suggests that Boltzmann machines
are more robust to noise.

One noticeable observation was that the GRBM outperformed,
in many cases, the DAE with two hidden layers which had
twice as many parameter. This potentially suggests that a
better inference of approximate posterior distribution over
the hidden units might make GDBMs outperform, or comparable
to, DAEs with the same number of hidden layers and units.
More work will be required in the future to make a definite
answer to this question.

Although it is difficult to make any general conclusion from
the experiments, it was evident that deep models, regardless
of whether they are DAEs or BMs, performed better and were
more robust to the level of noise than their more shallow
counterparts. In the future, it might be appealing to
investigate the possibility of combining multiple deep
neural networks with various depths to achieve better
denoising performance.



\bibliographystyle{plainnat}
\bibliography{denoising}

\begin{thebibliography}{27}
\providecommand{\natexlab}[1]{#1}
\providecommand{\url}[1]{\texttt{#1}}
\expandafter\ifx\csname urlstyle\endcsname\relax
  \providecommand{\doi}[1]{doi: #1}\else
  \providecommand{\doi}{doi: \begingroup \urlstyle{rm}\Url}\fi

\bibitem[Ackley et~al.(1985)Ackley, Hinton, and Sejnowski]{Ackley1985}
David~H. Ackley, Geoffrey~E. Hinton, and Terrence~J. Sejnowski.
\newblock {A learning algorithm for Boltzmann machines}.
\newblock \emph{Cognitive Science}, 9:\penalty0 147--169, 1985.

\bibitem[Burger et~al.(2012)Burger, Schuler, and Harmeling]{Burger2012}
H.C. Burger, C.J. Schuler, and S.~Harmeling.
\newblock Image denoising: Can plain neural networks compete with bm3d?
\newblock In \emph{Computer Vision and Pattern Recognition (CVPR), 2012 IEEE
  Conference on}, pages 2392 --2399, june 2012.

\bibitem[Cho et~al.(2013)Cho, Raiko, and Ilin]{Cho2013nc}
K.~Cho, T.~Raiko, and A.~Ilin.
\newblock Enhanced gradient for training restricted {B}oltzmann machines.
\newblock \emph{Neural Computation}, 2013.

\bibitem[Cho et~al.(2011{\natexlab{a}})Cho, Ilin, and Raiko]{Cho2011icann}
KyungHyun Cho, Alexander Ilin, and Tapani Raiko.
\newblock Improved learning of {G}aussian-{B}ernoulli restricted {B}oltzmann
  machines.
\newblock In \emph{Proc.\ of the 20th Int.\ Conf.\ on Artificial Neural
  Networks (ICANN 2010)}, 2011{\natexlab{a}}.

\bibitem[Cho et~al.(2011{\natexlab{b}})Cho, Raiko, and Ilin]{Cho2011dlufl}
KyungHyun Cho, Tapani Raiko, and Alexander Ilin.
\newblock {Gaussian-Bernoulli deep Boltzmann machine}.
\newblock In \emph{NIPS 2011 Workshop on Deep Learning and Unsupervised Feature
  Learning}, Sierra Nevada, Spain, December 2011{\natexlab{b}}.

\bibitem[Cho et~al.(2011{\natexlab{c}})Cho, Raiko, and Ilin]{Cho2011icml}
KyungHyun Cho, Tapani Raiko, and Alexander Ilin.
\newblock Enhanced gradient and adaptive learning rate for training restricted
  {B}oltzmann machines.
\newblock In \emph{Proc.\ of the 28th Int.\ Conf.\ on Machine Learning (ICML
  2011)}, pages 105--112, New York, NY, USA, June 2011{\natexlab{c}}. ACM.

\bibitem[Cho et~al.(2012)Cho, Raiko, Ilin, and Karhunen]{Cho2012dlufl}
KyungHyun Cho, Tapani Raiko, Alexander Ilin, and Juha Karhunen.
\newblock {A Two-Stage Pretraining Algorithm for Deep Boltzmann Machines}.
\newblock In \emph{NIPS 2012 Workshop on Deep Learning and Unsupervised Feature
  Learning}, Lake Tahoe, December 2012.

\bibitem[Dabov et~al.(2007)Dabov, Foi, Katkovnik, and Egiazarian]{Dabov2007}
Kostadin Dabov, Alessandro Foi, Vladimir Katkovnik, and Karen Egiazarian.
\newblock Image denoising by sparse {3-D} {Transform-Domain} collaborative
  filtering.
\newblock \emph{Image Processing, IEEE Transactions on}, 16\penalty0
  (8):\penalty0 2080--2095, August 2007.

\bibitem[Desjardins et~al.(2012)Desjardins, Courville, and
  Bengio]{Desjardins2012}
Guillaume Desjardins, Aaron~C. Courville, and Yoshua Bengio.
\newblock On training deep {B}oltzmann machines.
\newblock \emph{CoRR (Cornell Univ.\ Computing Research Repository)},
  abs/1203.4416, 2012.

\bibitem[Elad and Aharon(2006)]{Elad2006}
M.~Elad and M.~Aharon.
\newblock Image denoising via sparse and redundant representations over learned
  dictionaries.
\newblock \emph{Image Processing, IEEE Transactions on}, 15\penalty0
  (12):\penalty0 3736--3745, December 2006.

\bibitem[Hinton and Salakhutdinov(2006)]{Hinton2006}
G.~Hinton and R.~Salakhutdinov.
\newblock Reducing the dimensionality of data with neural networks.
\newblock \emph{Science}, 313\penalty0 (5786):\penalty0 504--507, July 2006.

\bibitem[Hinton(2002)]{Hinton2002}
Geoffrey Hinton.
\newblock Training products of experts by minimizing contrastive divergence.
\newblock \emph{Neural Computation}, 14:\penalty0 1771--1800, August 2002.

\bibitem[Hyv\"{a}rinen et~al.(1999)Hyv\"{a}rinen, Hoyer, and
  Oja]{Hyvarinen1999a}
Aapo Hyv\"{a}rinen, Patrik Hoyer, and Erkki Oja.
\newblock Image denoising by sparse code shrinkage.
\newblock In \emph{Intelligent Signal Processing}. IEEE Press, 1999.

\bibitem[Krizhevsky(2009)]{Krizhevsky2009}
A.~Krizhevsky.
\newblock {Learning multiple layers of features from tiny images}.
\newblock Technical report, Computer Science Department, University of Toronto,
  2009.

\bibitem[Lee et~al.(2008)Lee, Ekanadham, and Ng]{Lee2007}
Honglak Lee, Chaitanya Ekanadham, and Andrew Ng.
\newblock {Sparse deep belief net model for visual area V2}.
\newblock pages 873--880, 2008.

\bibitem[Lu et~al.(2011)Lu, Yuan, Yan, Li, and Li]{Lu2011}
Xiaoqiang Lu, Haoliang Yuan, Pingkun Yan, Luoqing Li, and Xuelong Li.
\newblock Image denoising via improved sparse coding.
\newblock In \emph{Proceedings of the British Machine Vision Conference}, pages
  74.1--74.0. BMVA Press, 2011.

\bibitem[Martin et~al.(2001)Martin, Fowlkes, Tal, and Malik]{Martin2001}
D.~Martin, C.~Fowlkes, D.~Tal, and J.~Malik.
\newblock A database of human segmented natural images and its application to
  evaluating segmentation algorithms and measuring ecological statistics.
\newblock In \emph{Proc. 8th Int'l Conf. Computer Vision}, volume~2, pages
  416--423, July 2001.

\bibitem[Neal and Hinton(1999)]{Neal1999}
Radford~M. Neal and Geoffrey~E. Hinton.
\newblock Learning in graphical models.
\newblock chapter A view of the EM algorithm that justifies incremental,
  sparse, and other variants, pages 355--368. MIT Press, Cambridge, MA, USA,
  1999.

\bibitem[Portilla et~al.(2003)Portilla, Strela, Wainwright, and
  Simoncelli]{Portilla2003}
J.~Portilla, V.~Strela, M.J. Wainwright, and E.P. Simoncelli.
\newblock Image denoising using scale mixtures of gaussians in the wavelet
  domain.
\newblock \emph{Image Processing, IEEE Transactions on}, 12\penalty0
  (11):\penalty0 1338 -- 1351, nov. 2003.

\bibitem[Salakhutdinov(2010)]{Salakhutdinov2010}
Ruslan Salakhutdinov.
\newblock Learning deep {B}oltzmann machines using adaptive {MCMC}.
\newblock In Johannes F{\"u}rnkranz and Thorsten Joachims, editors,
  \emph{Proc.\ of the 27th Int.\ Conf.\ on Machine Learning (ICML 2010)}, pages
  943--950, Haifa, Israel, June 2010. Omnipress.

\bibitem[Salakhutdinov and Hinton(2009)]{Salakhutdinov2009a}
Ruslan Salakhutdinov and Geoffrey~E. Hinton.
\newblock Deep {B}oltzmann machines.
\newblock In \emph{Proc.\ of the Int.\ Conf.\ on Artificial Intelligence and
  Statistics (AISTATS 2009)}, pages 448--455, 2009.

\bibitem[Shang and Huang(2005)]{Shang2005}
Li~Shang and Deshuang Huang.
\newblock Image denoising using non-negative sparse coding shrinkage algorithm.
\newblock In \emph{Computer Vision and Pattern Recognition, 2005. CVPR 2005.
  IEEE Computer Society Conference on}, volume~1, pages 1017 -- 1022 vol. 1,
  june 2005.

\bibitem[Smolensky(1986)]{Smolensky1986}
P.~Smolensky.
\newblock {Information processing in dynamical systems: foundations of harmony
  theory}.
\newblock In \emph{Parallel distributed processing: explorations in the
  microstructure of cognition, vol. 1: foundations}, pages 194--281. MIT Press,
  Cambridge, MA, USA, 1986.

\bibitem[Sonka et~al.(2007)Sonka, Hlavac, and Boyle]{Sonka2007}
Milan Sonka, Vaclav Hlavac, and Roger Boyle.
\newblock \emph{Image Processing, Analysis, and Machine Vision}.
\newblock Thomson-Engineering, 2007.
\newblock ISBN 049508252X.

\bibitem[Tieleman(2008)]{Tieleman2008}
Tijmen Tieleman.
\newblock \emph{{Training restricted Boltzmann machines using approximations to
  the likelihood gradient}}.
\newblock ICML '08. ACM, New York, NY, USA, 2008.

\bibitem[Vincent et~al.(2010)Vincent, Larochelle, Lajoie, Bengio, and
  Manzagol]{Vincent2010}
Pascal Vincent, Hugo Larochelle, Isabelle Lajoie, Yoshua Bengio, and
  Pierre-Antoine Manzagol.
\newblock Stacked denoising autoencoders: Learning useful representations in a
  deep network with a local denoising criterion.
\newblock \emph{Journal of Machine Learning Research}, 11:\penalty0 3371--3408,
  December 2010.

\bibitem[Xie et~al.(2012)Xie, Xu, and Chen]{Xie2012}
Junyuan Xie, Linli Xu, and Enhong Chen.
\newblock Image denoising and inpainting with deep neural networks.
\newblock In P.~Bartlett, F.C.N. Pereira, C.J.C. Burges, L.~Bottou, and K.Q.
  Weinberger, editors, \emph{Advances in Neural Information Processing Systems
  25}, pages 350--358. 2012.

\end{thebibliography}

\newpage
\appendix
\section{Training Procedures: Details}
\label{sec:training_detail}

Here, we describe the procedures used for training the deep
neural networks in the experiments.

\subsection{Data Preprocessing}

Prior to training a model, we normalized each pixel of the
training set such that, across all the training samples, the
mean and variance of each pixel are $0$ and $1$. 

The original mean and variance were discarded after
training. During test, we computed the mean
and variance of all image patches from each test image and
used them instead.

\subsection{Denoising Autoencoders}

A single-layer DAE was trained by the stochastic gradient
descent for 200 epochs. A minibatch of size 128 was used at
each update, and a single epoch was equivalent to one cycle
over all training samples. 

The initial learning rate was set to $\eta_0 = 0.05$ and was
decreased over training according to the following schedule:
\begin{align*}
    \eta_t = \frac{\eta_0}{1 + \frac{t}{5000}}.
\end{align*}

In order to encourage the sparsity of the hidden units, we
used the following regularization term
\begin{align*}
    -\lambda \sum_{n=1}^N \sum_{j=1}^q \left( \rho -
    \phi\left(\sum_{i=1}^{p} v_i^{(n)} w_{ij} + c_j\right)
    \right)^2,
\end{align*}
where $\lambda = 0.1$, $\rho = 0.1$ and $p$ and $q$ are
respectively the numbers of visible and hidden units. $\phi$
is a sigmoid function.

Before computing the gradient at each update, we added a
white Gaussian noise of standard deviation $0.1$ to all
components of an input sample and forced randomly chosen
20\% of input units to zeros.

The weights of a deep DAE was first initialized by
layer-wise pretraining. During the pretraining, each layer
was trained as if it were a single-layer DAE, following the
same procedure described above, except
that no white Gaussian noise was added for the layers other
than the first one.

After pretraining, we trained the deep DAE with the
stochastic backpropagation algorithm for 200 epochs using
minibatches of size 128. The initial learning rate was
chosen to be $\eta_0 = 0.01$ and the learning rate was
annealed according to
\begin{align*}
    \eta_t = \frac{\eta_0}{1 + \frac{t}{5000}}.
\end{align*}

For each denoising autoencoder regardless of its depth, we
used a tied set of weights for the encoder and decoder. 

\subsection{Restricted Boltzmann Machines}

We used the modified energy function of a Gaussian-Bernoulli
RBM (GRBM) proposed by \citet{Cho2011icann}, however, with a
single $\sigma^2$ shared across all the visible units.  Each
GRBM was trained for 200 epochs, and each update was
performed using a minibatch of size 128. 

A learning rate was automatically selected by the adaptive
learning rate \citep{Cho2011icann} with the initial learning
rate and the upper-bound fixed to $0.001$ and $0.001$,
respectively. After 180 epochs, we decreased the learning
rate according to
\[
\eta \leftarrow \frac{\eta}{t},
\]
where $t$ denotes the number of updates counted
\textit{after} 180 epochs of training.

A persistent contrastive divergence (PCD)
\citep{Tieleman2008} was used, and at each update, a single
Gibbs step was taken for the model samples. Together with
PCD, we used the enhanced gradient \citep{Cho2013nc},
instead of the standard gradient, at each update.

\subsection{Deep Boltzmann Machines}

We used the two-stage pretraining algorithm
\citep{Cho2012dlufl} to initialize the parameters of each
DBM. The pretraining algorithm consists of two separate
stages.

We utilized the already trained
single-layer and two-layer DAEs to compute the activations
of the hidden units in the even-numbered hidden layers of
GDBMs. No separate, further training was done for those
DAEs in the first stage.

In the second stage, the model was trained as an RBM using
the coupled adaptive simulated tempering
\citep[CAST,][]{Salakhutdinov2010} with the base inverse
temperature set of $0.9$ and $50$ intermediate chains
between the base and model distributions.  At least 50
updates were required to make a swap between the slow and
fast samples.

The initial learning rate was set to $\eta_0=0.01$ and the
learning rate was annealed according to
\[
\eta_t = \frac{\eta_0}{1 + \frac{t}{5000}}.
\]

Again, the modified form of an energy function
\citep{Cho2011dlufl} was used with a shared variance
$\sigma^2$ for all the visible units. However, in this case,
we did not use the enhanced gradient.

After pretraining, the GDBMs were further finetuned using
the stochastic gradient method together with the variational
approximation \citep{Salakhutdinov2009a}. The CAST was again
used with the same hyperparameters. The initial learning
rate was set to $0.0005$ and the learning rate was decreased
according to the same schedule used during the second stage.

\newpage
\section{Result Using a Training Set From Berkeley Segmentation
Dataset}
\label{sec:bsd500}

\begin{figure}[t]
    \centering
    \psfrag{DAE}[Bl][Bl][0.7][0]{DAE}
    \psfrag{sDAE2}[Bl][Bl][0.7][0]{DAE(2)}
    \psfrag{sDAE4}[Bl][Bl][0.7][0]{DAE(4)}
    \psfrag{WienerFilt}[Bl][Bl][0.7][0]{Wiener Filt}
    \psfrag{GRBM}[Bc][Bc][0.7][0]{GRBM}
    \psfrag{GDBM2}[Bl][Bl][0.7][0]{GDBM(2)}
    \psfrag{GDBM4}[Bl][Bl][0.7][0]{GDBM(4)}
    \psfrag{GRBM}[Bl][Bl][0.7][0]{GRBM}

    \psfrag{PSNR}[Bc][Bc][0.8][0]{PSNR}
    \psfrag{Noise Level}[Tc][Tc][0.8][0]{Noise Level}
    \psfrag{0.1}[Tc][Tc][0.8][0]{$0.1$}
    \psfrag{0.2}[Tc][Tc][0.8][0]{$0.2$}
    \psfrag{0.3}[Tc][Tc][0.8][0]{$0.3$}

    \begin{minipage}{0.1\textwidth}
        \centering
        $\phantom{\text{\textit{Aerials}}}$
    \end{minipage}
    \begin{minipage}{0.43\textwidth}
        \centering
        White noise
    \end{minipage}
    \begin{minipage}{0.43\textwidth}
        \centering
        Salt-and-pepper noise
    \end{minipage}
    \begin{minipage}{0.1\textwidth}
        \centering
        \textit{Aerials}
    \end{minipage}
    \begin{minipage}{0.43\textwidth}
        \centering
        \includegraphics[width=1\columnwidth]{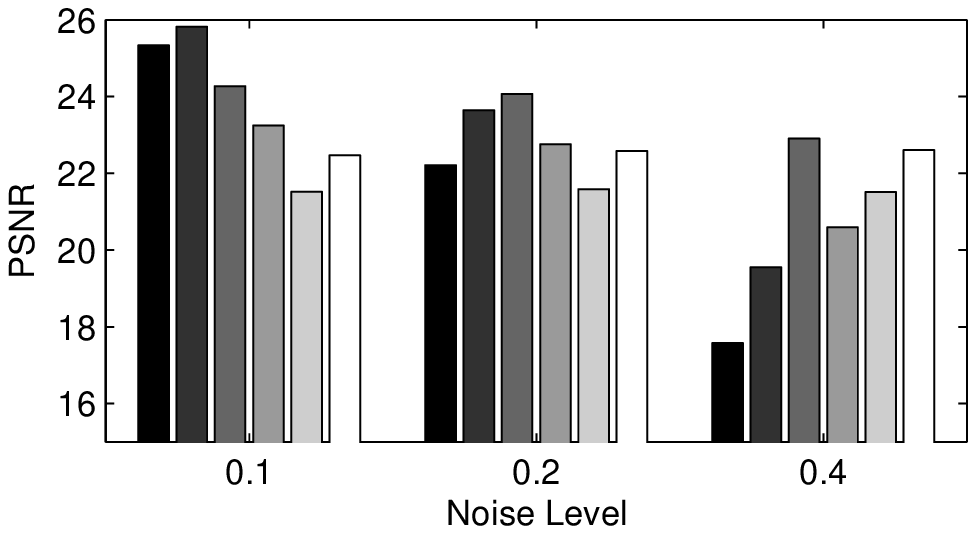}
    \end{minipage}
    \begin{minipage}{0.43\textwidth}
        \centering
        \includegraphics[width=1\columnwidth]{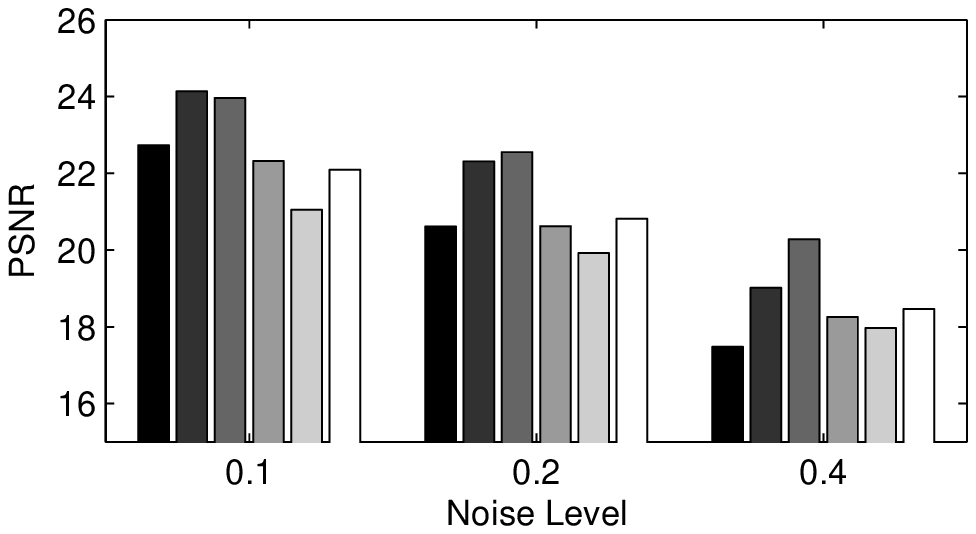}
    \end{minipage}
    \begin{minipage}{0.1\textwidth}
        \centering
        \textit{Textures}
    \end{minipage}
    \begin{minipage}{0.43\textwidth}
        \centering
        \includegraphics[width=1\columnwidth]{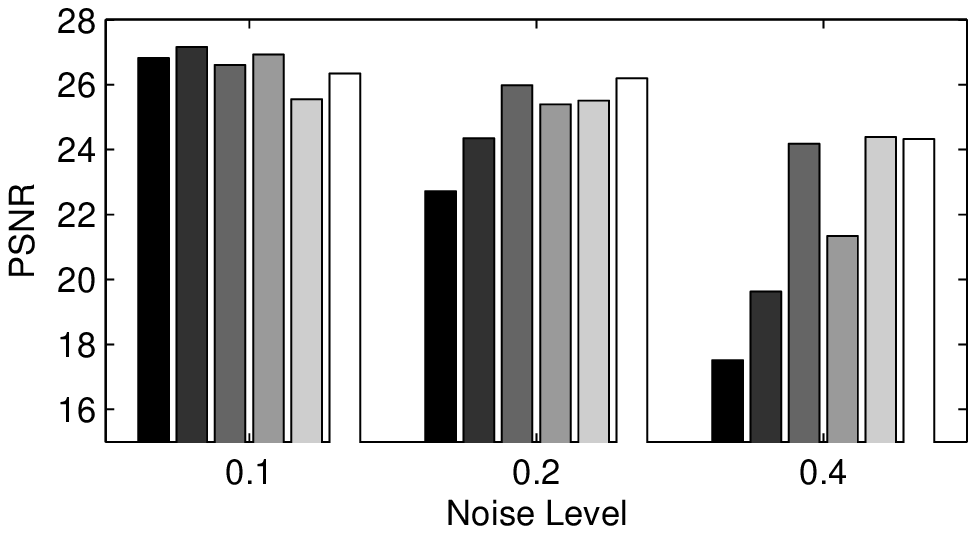}
    \end{minipage}
    \begin{minipage}{0.43\textwidth}
        \centering
        \includegraphics[width=1\columnwidth]{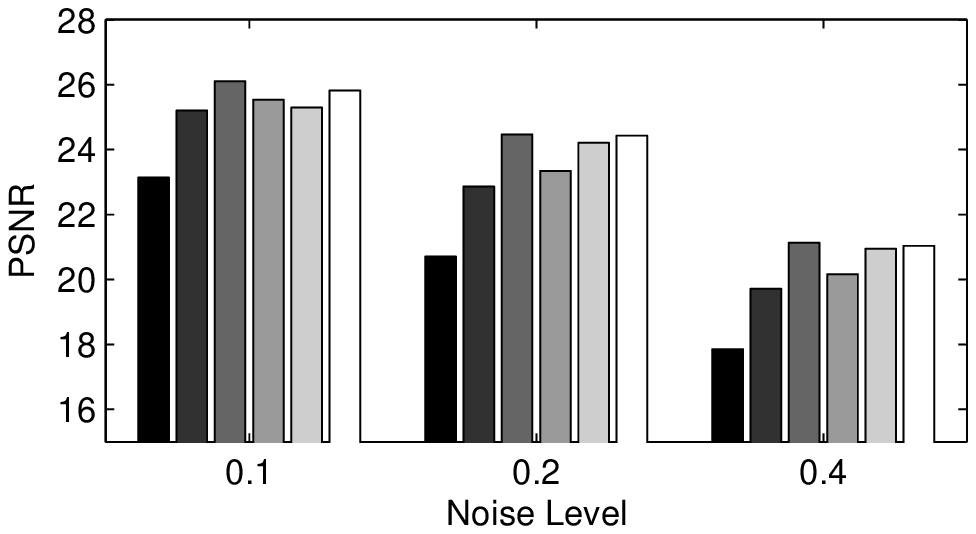}
    \end{minipage}
    \begin{minipage}{0.1\textwidth}
        \centering
        \textit{Misc.}
    \end{minipage}
    \begin{minipage}{0.43\textwidth}
        \centering
        \includegraphics[width=1\columnwidth]{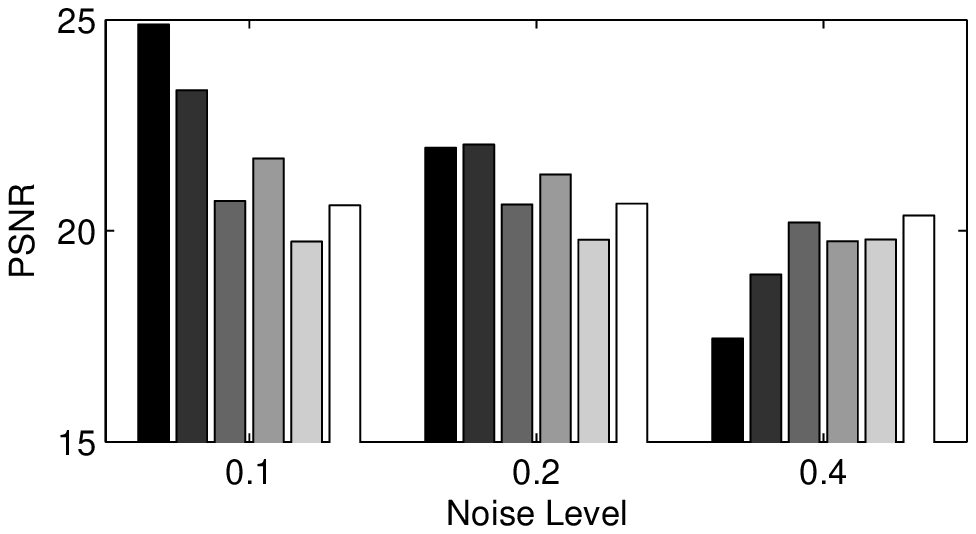}
    \end{minipage}
    \begin{minipage}{0.43\textwidth}
        \centering
        \includegraphics[width=1\columnwidth]{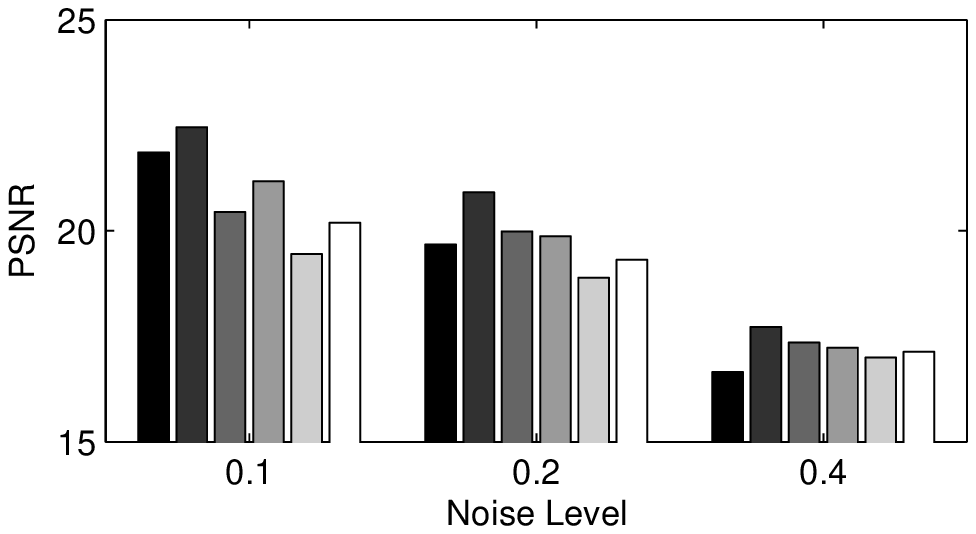}
    \end{minipage}
    \begin{minipage}{0.2\textwidth}
        ~
    \end{minipage}
    \begin{minipage}{0.76\textwidth}
        {
        \centering
        \includegraphics[width=1\columnwidth,clip=true,trim=0
        10 50 0]{legend.eps}
        }
    \end{minipage}

    \vspace{-6mm}
    \caption{PSNR of grayscale images corrupted by different
    types and levels of noise. The median PSNRs over the
    images in each set together. The models used for
    denoising in this case were trained on the training set
    constructed from the Berkeley Segmentation Dataset.}
    \label{fig:psnr_all_bsd500}
\end{figure}

Fig.~\ref{fig:psnr_all_bsd500} shows the result obtained by the
models trained on the training set constructed from the
Berkeley Segmentation Dataset. Although we see some minor
differences, the
overall trend is observed to be similar to that from the
experiment in the main text (see
Fig.~\ref{fig:psnr_all}).

Especially in the high-noise regime, the models with more
hidden layers tend to outperform those with only one or two
hidden layers. This agrees well with what we have observed
with the models trained on the training set constructed from
the CIFAR-10 dataset.

\end{document}